%% file: deep_continuous_networks.tex
\documentclass{article}
\usepackage{microtype}
\usepackage{graphicx}
\usepackage{subfigure}
\usepackage{booktabs} 

\usepackage{amsmath,amsfonts,bm}
\graphicspath{{figures/}}
\usepackage{wrapfig,booktabs}
\usepackage{multirow}
\usepackage{comment}
\usepackage[export]{adjustbox}

\usepackage{floatrow}
\newfloatcommand{capbtabbox}{table}[][\FBwidth]
\usepackage{makecell}
\usepackage{tabularx}
\newcommand{\model}{DCN}%
\usepackage{hyperref}

\usepackage[accepted]{icml2021}

\icmltitlerunning{Deep Continuous Networks}
\begin{document}

\twocolumn[
\icmltitle{Deep Continuous Networks}



\begin{icmlauthorlist}
\icmlauthor{Nergis Tomen}{Delft}
\icmlauthor{Silvia L. Pintea}{Delft}
\icmlauthor{Jan C. van Gemert}{Delft}
\end{icmlauthorlist}

\icmlaffiliation{Delft}{Computer Vision Lab, Delft University of Technology, Delft, Netherlands}

\icmlcorrespondingauthor{Nergis Tomen}{n.tomen@tudelft.nl}
\icmlkeywords{neuroscience, neural ODEs, Gaussian scale-space, learnable scale, receptive field size, continuous representations, pattern completion, computational neuroscience, convolutional neural networks}

\vskip 0.3in
]


\printAffiliationsAndNotice{}  

\begin{abstract}
CNNs and computational models of biological vision share some fundamental principles, which opened new avenues of research.
However, fruitful cross-field research is hampered by conventional CNN architectures being based on spatially and depthwise discrete representations, which cannot accommodate certain aspects of biological complexity such as continuously varying receptive field sizes and dynamics of neuronal responses.
Here we propose deep continuous networks (\model s), which combine spatially continuous filters, with the continuous depth framework of neural ODEs.
This allows us to learn the spatial support of the filters during training, as well as model the continuous evolution of feature maps, linking \model s closely to biological models.
We show that \model s are versatile and highly applicable to standard image classification and reconstruction problems, where they improve parameter and data efficiency, and allow for meta-parametrization.
We illustrate the biological plausibility of the scale distributions learned by \model s and explore their performance in a neuroscientifically inspired pattern completion task.
Finally, we investigate an efficient implementation of \model s by changing input contrast.
\end{abstract}

\section{Introduction}
Computational neuroscience and computer vision have a long and mutually beneficial history of cross-pollination of ideas~\citep{Cox2014,Sejnowski2020}. The current state-of-the-art in computer vision relies heavily on convolutional neural networks (CNNs), from which multiple analogies can be drawn to biological circuits~\citep{Kietzmann2019}. Thus, based on recent developments in deep learning, there has been a growing trend to relate CNNs to biological circuits~\citep{BrainScore}, and employ CNNs as models of biological vision~\citep{Zhuang2020}.
Specifically, recent advances in CNNs have enabled researchers to learn more accurate models of the response properties of neurons in the visual cortex~\citep{Klindt2017,Cadena2019,Ecker2019}, as well as to test decades old hypotheses from neuroscience in the domain of computer vision~\citep{Lindsey2019}. Hence, links between CNNs and biological models from neuroscience is fruitful for both research fields.

Contrary to many biological models, feed-forward CNNs typically use spatio-temporally discrete representations: CNNs employ spatially discretized, pixel-based kernels, and input is processed through a depthwise-discrete pipeline made up of successive convolutional layers.
To clarify, within our framework we consider CNN depth to be analogous to time, similar to input-processing time-course in biological models.
Unlike CNNs, large-scale, neuroscientific neural network models of the visual system often adopt continuous, closed-form expressions to describe spatio-temporal receptive fields, as well as the interaction strength between populations of neurons~\citep{Dayan2001}.
Among others, such descriptions serve to limit the scope and parameter space of a model, by utilizing prior information regarding receptive field shapes~\citep{Jones1987} and principles of perceptual grouping~\citep{Li1998}. In addition, the choice of continuous---and often analytic---functions help retain some analytical tractability in complex models involving a large number of coupled populations. Our approach draws inspiration from such computational models to propose continuous CNNs.

In this work we aim for a biologically more plausible CNN model: We bring together (a) spatially continuous receptive fields, where both the shape and the scale of the filters are trainable in the continuous domain, and (b) depthwise continuous representations capable of modeling the continuous evolution of neuronal responses in feed-forward CNNs.
Continuous receptive fields provide a link between modern CNNs and large-scale rate-based models of the visual system~\cite{Ernst2001}.
In addition, recent influential work in deep learning has introduced neural ordinary differential equations (ODEs)~\citep{Lu2018,Chen2018,Ruthotto2019} which propose a continuous depth (or time) interpretation of CNNs, while having spatially discrete filters. Such continuous depth models both offer end-to-end training capabilities with backpropagation which are highly applicable to computer vision problems (e.g. by way of adopting ResNet blocks~\citep{He2015}), as well as help bridge the gap to computational biology where networks are often modelled as dynamical systems which evolve according to differential equations.
Building on this, we introduce deep continuous networks (\model s), which are spatially and depthwise continuous in that the neurons have spatially well-defined receptive fields based on scale-spaces and Gaussian derivatives~\citep{florack1996gaussianLocalJet} and their activations evolve according to equations of motion comprising convolutional layers.
Thus, we combine depthwise and spatial continuity by employing neural ODEs in a network which learns linear weights for a set of analytic basis functions (as opposed to pixel-based weights), which can also intuitively be parametrized as a function of network depth (or time).

Our main contributions are:
(i)~We provide a theoretical formulation of deep networks with spatially and depthwise continuous representations, building on Gaussian basis functions and neural ODEs;
(ii)~We demonstrate the applicability of \model~models, namely, that they exhibit a reduction in parameters, improve data efficiency and can be used to parametrize convolutional filters as a function of depth in a straightforward fashion, while achieving performance comparable with or better than ResNet and ODE-Net baselines;
(iii)~We show that filter scales learned by \model s are consistent with biological observations and we propose that the combination of our design choices for spatial and depthwise continuity may be helpful in studying the emergence of biological receptive field properties as well as high-level phenomena such as pattern completion;
(iv)~We explore, for the first time, contrast sensitivity of neural ODEs and suggest that the continuous representations learned by \model s may be leveraged for computational savings.

We believe \model s can bring together two communities as they both provide a test bed for hypotheses and predictions pertaining to biological systems, and push the boundaries of biologically inspired computer vision.

\input{methods}
\input{results}

\input{related}

\input{discussion}

\bibliography{deep_continuous_networks}
\bibliographystyle{icml2021}


%



\end{document}


\begin{center}
    \textbf{\Large Deep Continuous Networks - Supplementary Material} \\[\baselineskip]
    \textbf{\Large ICML 2021 Submission} \\[\baselineskip]
    \textbf{\Large Paper ID: 3234}
\end{center}

\input{appendix}

\pagebreak

\bibliographystyle{icml2021}
\bibliography{deep_continuous_networks}

%% file: methods.tex
\section{Deep Continuous Networks}
\subsection{Neuroscientific motivation}
There is little doubt that modern deep learning frameworks will be conducive to effective and insightful collaborations between neuroscience and machine learning~\citep{Richards2019}.
In particular in vision research, CNNs are becoming increasingly popular for modelling early visual areas \citep{Batty2017,Ecker2019,Lindsey2019}.
Here we propose the \model~model which can facilitate such investigations by linking the end-to-end trainable but discrete CNN architectures with the spatio-temporally continuous models of biological vision. Our approach makes it possible to optimize the spatial extent (kernel size) of the filters during training, as well as explicitly model the dynamics of the neuronal responses to input images.

\textbf{Structured receptive fields.}
Classical receptive fields (RFs) of cortical neurons display complex response properties with a wide array of selectivity structures already at early visual areas~\citep{VandenBergh2010}. Such response properties may also vary greatly based on multiple factors. For example the RF size (spatial extent) is known to depend on eccentricity~\citep{Harvey2011}, visual area~\citep{Smith2001} and even cortical layer~\citep{Bauer1999}.
Similarly, studies have shown that spatial frequency selectivity and receptive field size  may co-vary with input contrast~\citep{Sceniak2002}.

Based on these observations, we aim to build a model which can accommodate the biological realism better than conventional CNNs, by explicitly modelling the RF size as a trainable parameter.
To that end, we adopt a Gaussian scale-space representation for the convolutional filters, which we call structured receptive fields (SRFs)~\citep{jacobsen2016structured}.
Previously, Gaussian scale-spaces have been proposed as a plausible model of biological receptive fields and feature extraction in low-level vision~\citep{florackIVC92scaleAndDiffStruct,Lindeberg1993,Lindeberg1994}. Here, we are inspired by computational models which investigate the origin of response properties in the visual system, by employing RFs and recurrent interaction functions which scale as a difference of Gaussians~\citep{Somers1995,Ernst2001}.

\begin{figure*}[ht]
    \hspace{0.4cm}
    \includegraphics[width=0.9\textwidth]{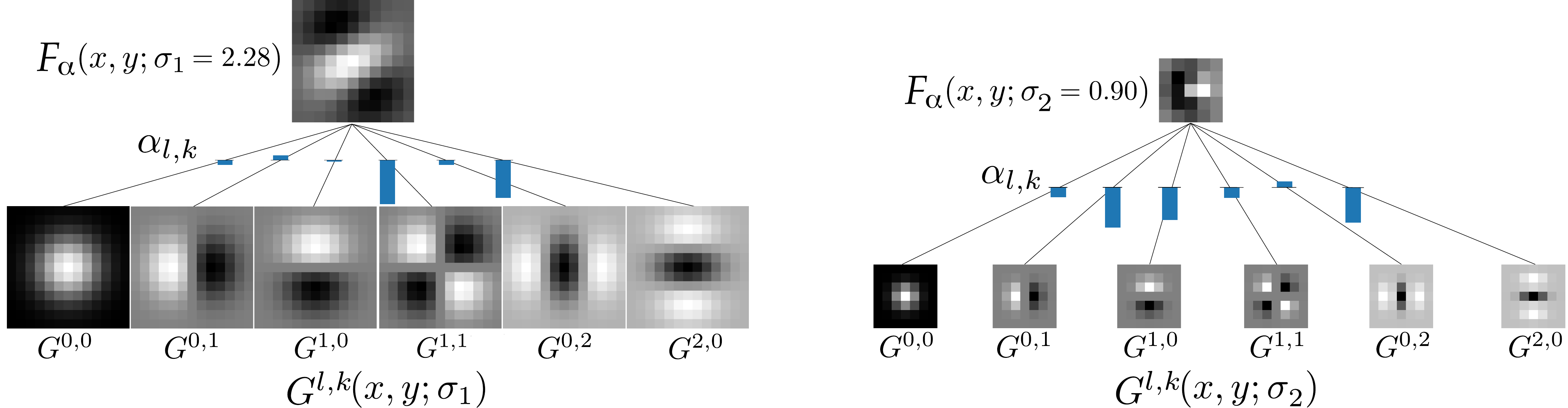}
    \vspace{-0.1cm}
    \caption{SRF filters based on N-jet filter approximation. Convolutional filters are defined as the weighted sum of Gaussian derivative basis functions up to order 2, with corresponding scales ${\sigma_1 = 2.28}$ (left) and ${\sigma_2=0.90}$ (right).
    Our \model~models learn both the coefficients $\alpha$, and the scale~$\sigma$ end-to-end during training.\vspace{-0.02cm}}
\label{fig:njet}
\end{figure*}

\textbf{Neural ODEs.} Studies have shown that both the contrast~\citep{Albrecht2002} and spatial frequency~\citep{Frazor2004} response functions of cortical neurons display characteristic temporal profiles.
However, temporal dynamics are not incorporated into standard feed-forward CNN models.
In addition, it has been suggested that lateral interactions play an important role in the generation of complex and selective neuronal responses~\citep{Angelucci2006}. Such activity dynamics are often computationally modeled using recurrently coupled neuronal populations whose activations evolve according to coupled differential equations~\citep{BenYishai1995,Ernst2001}.

To describe the continuous evolution of feature maps consistent with biological models, we adopt the framework of neural ODEs~\citep{Chen2018}.
Neural ODE interpretation of ResNet models presents an opportunity to explicitly model the dynamics of feature extractors in feed-forward CNNs. Under certain assumptions, neural ODEs can be interpreted as biologically plausible recurrent interactions~\citep{Liao2016, rousseau2019residual}, where the depth dimension represents time. Unlike neural ODEs with pixel-based convolutional filters, \model s with structured filters (SRFs) also provide an intuitive way to parametrize the evolution of the kernels as a function of depth.

\textbf{Deep Continuous Networks.} \model s presented here combine structured receptive fields with neural ODEs. We view spatio-temporally continuous representations in end-to-end trainable networks as a link between modern CNN architectures and computational models of biological vision. Specifically, we are inspired by large-scale models of population activity. In contrast, networks modelling biological phenomena at smaller spatio-temporal scales may require discrete descriptions of biological neurons, such as spatially-discrete photoreceptors or temporally-discrete spiking dynamics. However, continuous rate-based population models provide reasonably good explanations of phenomena observed at the network or systems level~\citep{BenYishai1995,Dayan2001}, which we believe align well with CNNs trained on high-level computer vision tasks.

Taken together, \model s provide a fully trainable analog to biological models with continuous receptive fields and continuously evolving state variables, while preserving the modularity of the visual hierarchy by stacking spatio-temporally continuous blocks in a feed-forward stream (Fig.~\ref{fig:architecture}).

\subsection{Structured receptive fields}
We use the multiscale local N-jet formulation~\citep{florack1996gaussianLocalJet} to define the filters in convolutional layers. Structured receptive fields (SRFs) based on the Gaussian N-jet basis functions are highly applicable to CNNs, as they represent a Taylor expansion of the input image or feature maps in a local neighbourhood in space and scale, and can be used to approximate pixel-based filters (Appendix~A.1). This means that each filter $F(x,y;\sigma)$ in the network is a weighted sum of $N$ basis functions, which are partial derivatives of the isotropic two-dimensional Gaussian function ${G(x,y; \sigma) =  \frac{1}{2 \pi \sigma^2} e^{\frac{-(x^2+y^2)}{2\sigma^2}}}$. The scale, or the spatial extent, of the filter is explicitly modelled in the $\sigma$ parameter of the Gaussian, which also indirectly determines the spatial frequency response of the SRF (Fig.~\ref{fig:njet}). Note that the Gaussian SRF formulation allows for learning filters with different aspect ratios, however, in this work we only consider isotropic basis functions with $\sigma=\sigma_x=\sigma_y$.

The N-jet formulation of an SRF filter $F(x,y)$ is given by:
\vspace{-0.2cm}
\begin{align}
  F_\alpha(x, y; \sigma)\ &=\ \sum_{\substack{0\ \le\ l,\ 0\ \le\ k}}^{l + k\ \le\ N}\ 
  \alpha_{l, k}\ G^{l,k}\left(x, y;\ \sigma \right) \nonumber \\[1.3ex]
  &=\sum_{\substack{0\ \le\ l,\ 0\ \le\ k}}^{l + k\ \le\ N}\ \alpha_{l, k}\ \frac{\partial^{l+k}}{\partial x^{l} \partial y^{k}} G\left(x, y;\ \sigma \right),
\label{eq:taylor2}
\end{align}
where $G^{l,k}\left(x, y;\ \sigma \right)$ are the partial derivatives of the Gaussian $G(x, y;\sigma)$ with respect to $x$ and $y$, $N$ is the degree of the Taylor polynomial which determines the basis order, and $\alpha$ encodes the expansion coefficients.

N-jet SRFs have favourable properties over pixel-based filters. SRF filters are steerable by the coefficients $\alpha$ and the basis functions are spatially separable. Likewise, due to their spatially continuous description, the filters can be trivially scaled, or rotated, without interpolation. In addition, SRFs can provide parameter efficiency when filters are constructed using a small number of basis functions. In this work we opt for basis order 2 (basis function up to the second order derivative), which yields relatively smooth filters. However, the generalized SRF framework allows for learning more irregular RF shapes by increasing the number of basis functions.

Fig.~\ref{fig:njet} shows the N-jet approximation of two filters in different scales $\sigma_1$ and $\sigma_2$. 
We note that both the coefficients $\alpha$ and the scale $\sigma$ are learnable filter parameters. 
Instead of fixing  the scale~$\sigma$ \emph{a~priori} and optimizing for $\alpha$ as in \citet{jacobsen2016structured} and \citet{sosnovik2019scale}, we integrate both these parameters in the network optimization, thus learning not only the shape but also the spatial support of the filters.

\subsection{Neural ODEs}
We model the continuous evolution of feature maps as a function of depth $t$ within an `ODE block'.
Formally, an ODE block contains a stack of $M$ convolutional layers, each with its own convolutional filters \textbf{w}$_m$ with $m=1 \ldots M$, followed by normalization G$_\textrm{norm}(\cdot)$ and non-linear activation CELU$(\cdot)$ functions. Following the notations of~\citet{Chen2018} and~\citet{Ruthotto2019}, we define the equations of motion for the feature states~$\mathbf{h} \in \mathbb{R}^n$ as:
\begin{align} \label{eq:EoM}
    &\frac{d\mathbf{h}(t)}{dt}
    =f(\mathbf{h}(t),t,\textbf{w}_{m},\textbf{d}_{m}) \\
    &\,\,\,=\textrm{G$_\textrm{norm}$}\left[\textbf{K}_2(\textbf{w}_2) g(\textbf{K}_1(\textbf{w}_1)g(\textbf{h})+\textbf{d}_1 t)+\textbf{d}_2 t\right] \nonumber
\end{align}
where $g(\textbf{x})=\textrm{CELU}(\textrm{G$_\textrm{norm}$}(\textbf{x}))$, the linear operators ${\textbf{K}_m(\cdot) \in \mathbb{R}^{n \times n}}$ denote the convolution operators parametrized by \textbf{w}$_m$. The filters \textbf{w}$_m(\theta)$, \textbf{d}$_m(\theta)$ are functions of learnable parameters $\theta$. In conventional CNNs \textbf{w}$_m$ are typically $3 \times 3$ kernels where the learnable parameters correspond to pixel weights. In the \model~model we define the filters \textbf{w}$_m$ using the Gaussian SRF, thus, the learnable parameters are basis coefficients $\alpha$ and scale $\sigma$, and the kernel size is not fixed but scales with $\sigma$ and is learned (see Section~\ref{ssec:dcn}).

The CNN convolution operator, $\textbf{K}_m(\textbf{w}_m)$ with 2 input and 2 output channels can be written as
\begin{equation}
    \textbf{K}_m(\textbf{w}_m) = \begin{pmatrix}
\textbf{K}_m^{1,1}(\textbf{w}_m^{1,1}) & \textbf{K}_m^{1,2}(\textbf{w}_m^{1,2})\\ 
\textbf{K}_m^{2,1}(\textbf{w}_m^{2,1}) & \textbf{K}_m^{2,2}(\textbf{w}_m^{2,2})
\end{pmatrix}
\end{equation}
with $\textbf{w}_{m}^{ji}$ the convolutional kernels for input channel $i$ and output channel $j$ of the $m$-th convolution. The time-offset terms \textbf{d}$_m t$ in Eq.~\ref{eq:EoM} makes the ODE an explicit function of $t$, which separates the ODE block implementation from a simple convolutional block with weight sharing over depth.

In accordance with conventional ResNet blocks, we pick $M=2$.
Based on the implementation by~\citet{Chen2018}, $G_\textrm{norm}$ is defined as group normalization~\citep{Wu2018}.
For generalized compatibility with ODEs and the adjoint method, we choose a non-linear activation function with a theoretically unique and bounded adjoint, namely continuously differentiable exponential linear units, or CELU~\citep{Barron2017}.
Similarly, we keep the linear dependence of the equations of motion on continuous network depth $t$.
Finally, we adapt the GPU implementation of ODE solvers\footnote{https://github.com/rtqichen/torchdiffeq/} to solve the equations of motion for a predefined time interval $t \in [0,T]$ using the adaptive step size DOPRI method.

\begin{figure}
    \includegraphics[trim=0 0 0 0,clip,width=0.9\textwidth]{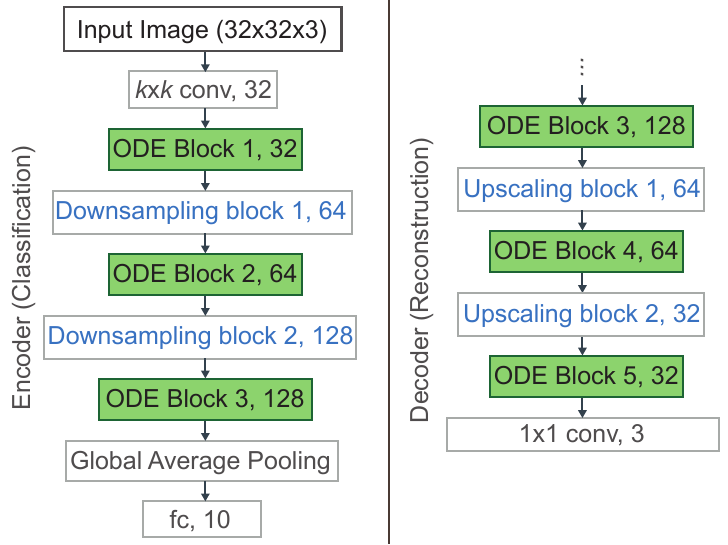}
    \vspace{-0.4cm}
    \caption{\model~model architecture with CIFAR-10 input images. Convolutional kernel size $k$ is learned during training. The equations of motion (Eq.~\ref{eq:EoM}) are solved within ODE blocks.\vspace{-0.1cm}}
    \label{fig:architecture}
\end{figure}

\textbf{Time vs. depth} In the neural ODE definition~\cite{Chen2018}, the discrete depth of feed-forward networks such as ResNets is reimagined as a continuous dimension denoted by time $t$, where the input image defines the initial conditions $\mathbf{h}(0)$. For the rest of this paper, we use the interpretation that the number of function evaluations performed by the numerical ODE solver is analogous to network depth. In this sense, continuous `depth' or `time' refers to the continuous variable $t$ within the ODE blocks, while the full architecture is still modular and composed of multiple ODE blocks. It is also important to note that when we talk about the spatio-temporal dynamics of DCNs, we refer to the temporal evolution of the feature maps in the ODE blocks and not to input dynamics, as in a video. While \model s are primarily feed-forward networks, the ODE definition makes it possible for DCNs to model time-varying neuronal activations via the `continuous depth', even in response to static input images (see Section~\ref{sec:related} for more detailed comparisons with recurrent neural networks).

\subsection{Deep Continuous Networks with SRFs and ODEs}
\label{ssec:dcn}
We formulate deep continuous networks (\model s) by employing learnable, continuous SRF filter descriptions to define the weights $\textbf{w}$ in the evolution of a neural ODE. This means that for \model s, each $\textbf{w}_{m}^{ji}$ in Eq.~\ref{eq:EoM}
is a discretization of the continuous SRF filter $F_{\alpha_m^{ji}}(x,y;\sigma_m)$ given in Eq.~\ref{eq:taylor2}, sampled in $[-2\sigma_m,2\sigma_m]$. $\alpha_m^{ji}$ and $\sigma_m$ are trainable filter parameters, where $\sigma_m^{ji} = \sigma_m$ is shared between the filters in a convolutional layer $m$ unless stated otherwise. All our code is available at\footnote{https://github.com/ntomen/Deep-Continuous-Networks}.

\textbf{Network architecture and training.} We construct \model s by stacking ODE blocks separated by downsampling blocks (Fig.~\ref{fig:architecture}). Each downsampling block is a sequence of normalization, nonlinear activation and strided convolution. We use a convolutional layer for increasing the channel dimensionality at the input level and employ global average pooling and a fully connected layer at the output level. We train our networks using cross-entropy loss and the CIFAR-10 dataset~\citep{cifar10}. (See Appendices~A.2-A.3 for further details regarding training parameters.)

\begin{table}
    \centering
    \small
    \setlength\tabcolsep{3.8pt} 
    \caption{CIFAR-10 validation accuracies of \model~models, averaged over 3 runs, compared to baseline models. ODE-Net and ResNet-blocks baselines are as introduced in *\citet{Chen2018}. \model s perform on par with spatially and\slash or temporally discrete baselines, despite having a lower number of trainable parameters.}
    \label{tb:accuracies}
    \begin{tabular}{lcccc}
        \toprule
        \multirow{2}{*}{Model} &  \multicolumn{2}{c}{Continuity} & \multirow{2}{*}{\makecell{\vspace{-0.25cm}\\ Accuracy \\ (\%)}}  & \multirow{2}{*}{\makecell{\vspace{-0.25cm}\\ Para- \\ meters}} \\
        \cmidrule{2-3}
        {} & Spatial & Temporal & {} & {} \\
        \midrule
        ODE-Net * & \text{\sffamily x} & \checkmark & 89.6 $\pm$ 0.3  & 560K \\
        ResNet-blocks * & \text{\sffamily x} & \text{\sffamily x} & 89.0 $\pm$ 0.2 & 555K \\
        ResNet-SRF-blocks & \checkmark & \text{\sffamily x} & 88.3 $\pm$ 0.03 & 426K \\
        ResNet-SRF-full & \checkmark & \text{\sffamily x} & 89.3 $\pm$ 0.4 & 323K \\
        \midrule
        \model-ODE & \checkmark & \checkmark & 89.5 $\pm$ 0.2 & 429K  \\
        \model-full & \checkmark & \checkmark & 89.2 $\pm$ 0.3 & 326K \\
        \model~$\sigma^{ji}$ & \checkmark & \checkmark & 89.7 $\pm$ 0.3 & 472K \\
        \bottomrule
        \vspace{-0.3cm}
    \end{tabular}
\end{table}

\begin{table*}
    \caption{Validation accuracies for the \model-ODE model and baselines trained on a subset of CIFAR-10 (small-data regime). First two rows show small-data baseline accuracies taken from $^{\dagger}$\citet{Arora2020}. ResNet-blocks and ODE-Net models are implemented by us, as in Table~\ref{tb:accuracies}. The \model~model outperforms spatially and\slash or temporally discrete baselines for medium training set size as parameter efficiency leads to data efficiency. All results are averaged over 3 runs.}
    \label{tb:small_data}
    \hspace{-1.8cm}
    \setlength\tabcolsep{2.3pt} 
    \fontsize{8.6}{5}\selectfont
    \begin{tabularx}{0.9\textwidth}{lc|c|c|c|c|c|c|c|c|c|c}
        \cmidrule[0.8pt]{1-12} \\[-1.1ex]
        \multirow{2}{*}{Model} &  \multicolumn{11}{c}{\# images per class} \\
        \cmidrule{2-12}
        {} & 2 & 4 & 8 & 16 & 32 & 52 & 64 & 103 & 128 & 512 & 1024 \\
        \cmidrule{1-12} \\[-1.0ex]
        ResNet34$^{\dagger}$ & 17.5 $\!\pm\!$ 2.5 & 19.5 $\!\pm\!$ 1.4 & 23.3 $\!\pm\!$ 1.6 & 28.3 $\!\pm\!$ 1.4 & 33.2 $\!\pm\!$ 1.2 & {--} & 41.7 $\!\pm\!$ 1.1 & {--} & 49.1 $\!\pm\!$ 1.3 & {--} & {--}\\[0.6ex]
        CNTK$^{\dagger}$ & \textbf{18.8} $\!\pm\!$ 2.1 & \textbf{21.3} $\!\pm\!$ 1.9 & 25.5 $\!\pm\!$ 1.9 & 30.5 $\!\pm\!$ 1.2 & 36.6 $\!\pm\!$ 0.9 & {--} & 42.6 $\!\pm\!$ 0.7 & {--} & 48.9 $\!\pm\!$ 0.7 & {--} & {--} \\[0.6ex]
        \cmidrule{1-12} \\[-0.4ex]
        ResNet-blocks & 16.7 $\!\pm\!$ 0.8 & 19.6 $\!\pm\!$	1.0 & 22.0	$\!\pm\!$ 1.3 & 28.1	$\!\pm\!$	1.7 & 35.4	$\!\pm\!$	0.9 & 39.8 $\!\pm\!$ 0.6 & 41.6	$\!\pm\!$	1.5 & 49.0 $\!\pm\!$ 0.2 & 50.9	$\!\pm\!$	0.6 & 70.4 $\!\pm\!$ 1.2 & 76.8 $\!\pm\!$ 0.7 \\[1.2ex]
        ODE-Net & 16.8 $\!\pm\!$ 2.8 & 20.5 $\!\pm\!$ 0.8 & 23.1 $\!\pm\!$ 2.5 & 29.8 $\!\pm\!$	0.8 & 36.4 $\!\pm\!$	1.0 & 41.7 $\!\pm\!$ 1.2 & 42.3	$\!\pm\!$ 0.2 & 48.6 $\!\pm\!$ 0.5 & 50.7	$\!\pm\!$ 0.7 & 71.7 $\!\pm\!$ 1.5 & 77.4 $\!\pm\!$ 0.5 \\[1.2ex]
        \model-ODE & 16.4 $\!\pm\!$ 1.6 & 19.8	$\!\pm\!$	0.7 & \textbf{26.5}	$\!\pm\!$	0.9 & \textbf{31.2}	$\!\pm\!$	0.6 & \textbf{37.7}	$\!\pm\!$	0.6 & \textbf{44.5} $\!\pm\!$ 0.8 & \textbf{48.0}	$\!\pm\!$	1.3 & \textbf{54.2} $\!\pm\!$ 0.8 & \textbf{58.2}	$\!\pm\!$	0.7 & \textbf{75.5} $\!\pm\!$ 0.8 & \textbf{79.7} $\!\pm\!$ 0.3 \\[0.6ex]
        \cmidrule[0.8pt]{1-12}
        \vspace{-0.6cm}
    \end{tabularx}
\end{table*}

As a baseline without spatial continuity, we compare \model~performance to the ODE-Net introduced in~\citet{Chen2018}, where the convolutions within the ODE blocks are performed using discrete, pixel-based kernels, with $3 \times 3$ parameters. As a baseline without (depthwise) temporal continuity, we define the `ResNet-blocks' model where the ODE blocks are replaced by generic, discrete ResNet blocks, comprising two convolutional layers and a skip connection, with comparable number of parameters to the ODE-Net. This is also a baseline model used in~\citet{Chen2018}. In the ResNet-SRF-blocks model, we provide the discrete-depth and continuous-space baseline by replacing the $3 \times 3$ filter definition of ResNet-blocks with SRF definitions.

We test two versions of \model s and ResNet-SRF-blocks to quantify the viability of SRF filters outside of the ODE blocks. In \model-ODE and ResNet-SRF-blocks we use the SRF filters only within the ODE (ResNet) blocks, and for the remaining layers we use discrete kernels with the same hyperparameters as the baselines. In the second version, \model-full (ResNet-SRF-full), we use spatially continuous kernels everywhere, including the downsampling layers.

As an additional demonstration of the versatility of \model s, we conduct an image reconstruction experiment on CIFAR-10. We use the feature maps generated by encoder networks (output of ODE Block 3 in Fig.~\ref{fig:architecture}), as input to a decoder network. The decoder networks are composed of 2 \model-ODE, ODE-Net or ResNet blocks, separated by bilinear upscaling layers and $1 \times 1$ convolutions to reduce dimensionality. 

Finally, we investigate the case where we drop scale sharing within a layer, and optimize the scale parameter $\sigma^{ji}_m$ independently for each input channel $i$ and output channel $j$, which we call \model~$\sigma^{ji}$.

\textbf{Meta-parametrization.}
\model s enable us to parametrize the trainable filter parameters $\alpha$ and $\sigma$ as a function of depth $t$. This both enables the kernels to vary smoothly over depth, and lets us define temporal dynamics for the neuronal responses in our network. We test the viability of such models by introducing \model~variants where $\sigma$ and/or $\alpha$ are defined using linear or quadratic functions of $t$ and learnable parameters $a$, $b$, $c$, $a_s$, $b_s$, $a_{\alpha}$ and $b_{\alpha}$ (Table~\ref{tb:meta-def}).



%% file: results.tex
\section{Experimental Analysis}
\subsection{Parameter reduction and data efficiency}
Similar to biological models, where analytical receptive fields limit the scope of the model using prior information, we find that \model s are more parameter efficient compared to baseline networks.
Evaluated on CIFAR-10, \model s perform on par with baselines, despite using SRFs of a small basis order 2, which means each filter shape is defined by only 6 free parameters as opposed to 9 for the conventional $3 \times 3$ kernels (Table~\ref{tb:accuracies}).
In addition, we find that parameter reduction via the use of SRFs with a small basis order also leads to data efficiency. When trained on a subset of CIFAR-10 images (small-data regime), \model s outperform the discrete baseline networks (Table~\ref{tb:small_data}). We also find accuracy increase over the small-data performance reported by~\citet{Arora2020} for the convolutional neural tangent kernel (CNTK) model (Table~\ref{tb:small_data}).

Moreover, we train encoder-decoder networks to reconstruct CIFAR-10 images using mean squared error (MSE) loss. We find that the \model~models outperform discrete baseline models on the validation set (Table~\ref{tb:reconstruction}), despite having a lower number of parameters as before. Additional details and example images are shown in Appendix~A.4.

We find that meta-parametrized \model~variants match the classification performance of baselines and may outperform \model s with static weights (Table~\ref{tb:meta-accuracy}). This is an interesting finding as we test only a few models, with little hyperparameter optimization, indicating that \model s can potentially be used to parametrize the dependence of convolutional kernel weights on network depth, for further parameter reduction.

\begin{table}
    \caption{\model s achieve lower MSE loss in the reconstruction task than discrete baselines on the CIFAR-10 validation set, despite using a smaller number of parameters. See also Appendix A.4 for reconstructed image examples. All results are averaged over 3 runs.}
    \label{tb:reconstruction}
    \centering
    \setlength\tabcolsep{6pt} 
    \begin{tabularx}{0.6\textwidth}{lc}
        \cmidrule[0.8pt]{1-2}
        \multirow{2}{*}{Model} & \multirow{2}{*}{\makecell{\vspace{-0.35cm}\\ Reconstruction \\ Loss (\%)}} \\
        {} & {} \\
        \cmidrule{1-2}
        ResNet-blocks & 21.0 $\pm$ 0.4  \\
        ODE-Net & 20.2 $\pm$ 1.3\\
        \model-ODE & \textbf{17.1} $\pm$ 0.3 \\
        \cmidrule[0.8pt]{1-2}
        \vspace{-0.8cm}
    \end{tabularx}
\end{table}

\subsection{Link with biological models}
\begin{figure*}
    \vspace{0.5cm}
    \begin{tabular}{ccc}
        \hspace{0.4cm}
        \begin{tabular}{c}
            \includegraphics[trim=100 20 0 90, width=0.27\linewidth]{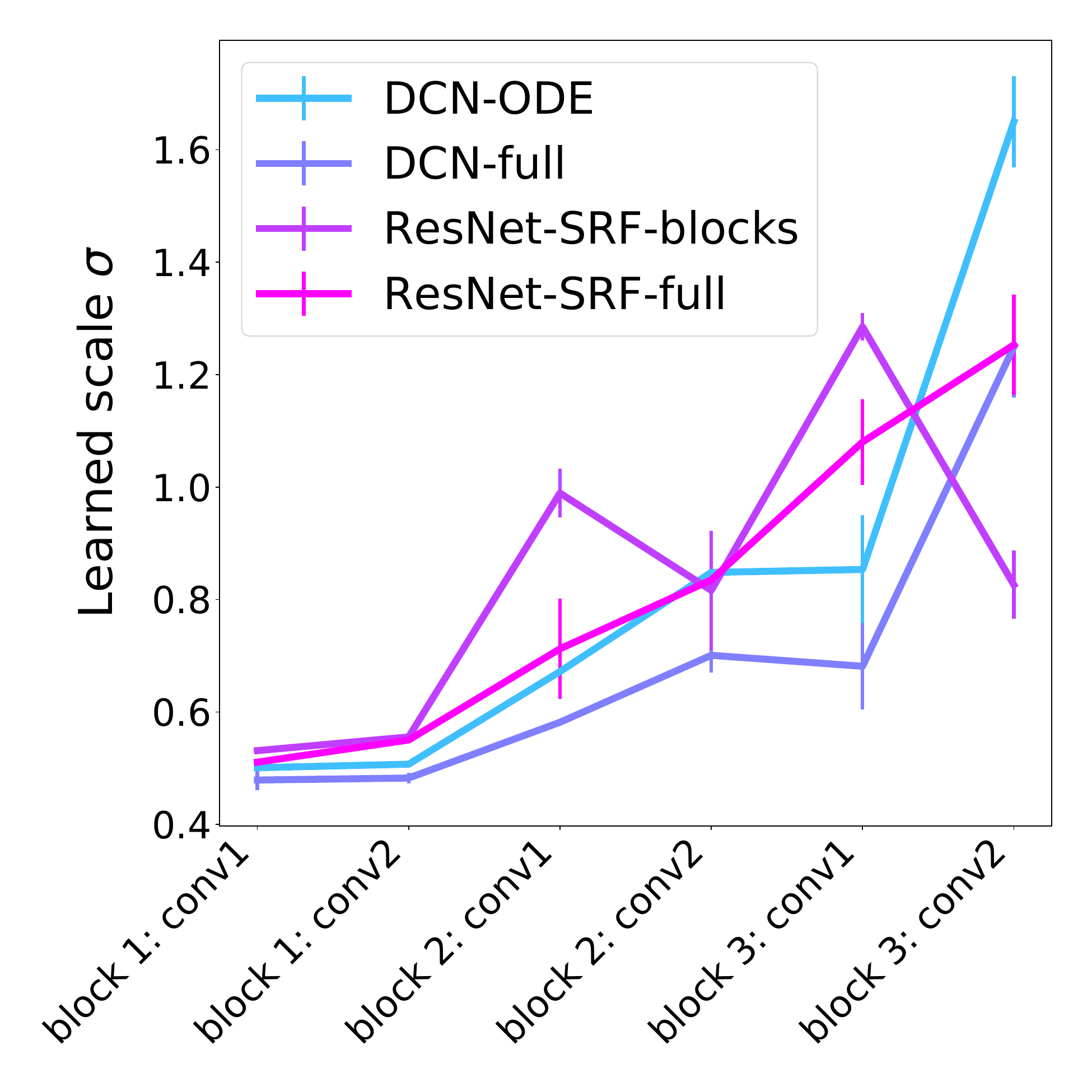}
        \end{tabular}
        &
        \hspace{0.3cm}
        \begin{tabular}{c}
            \includegraphics[trim=130 20 0 230, width=0.236\linewidth]{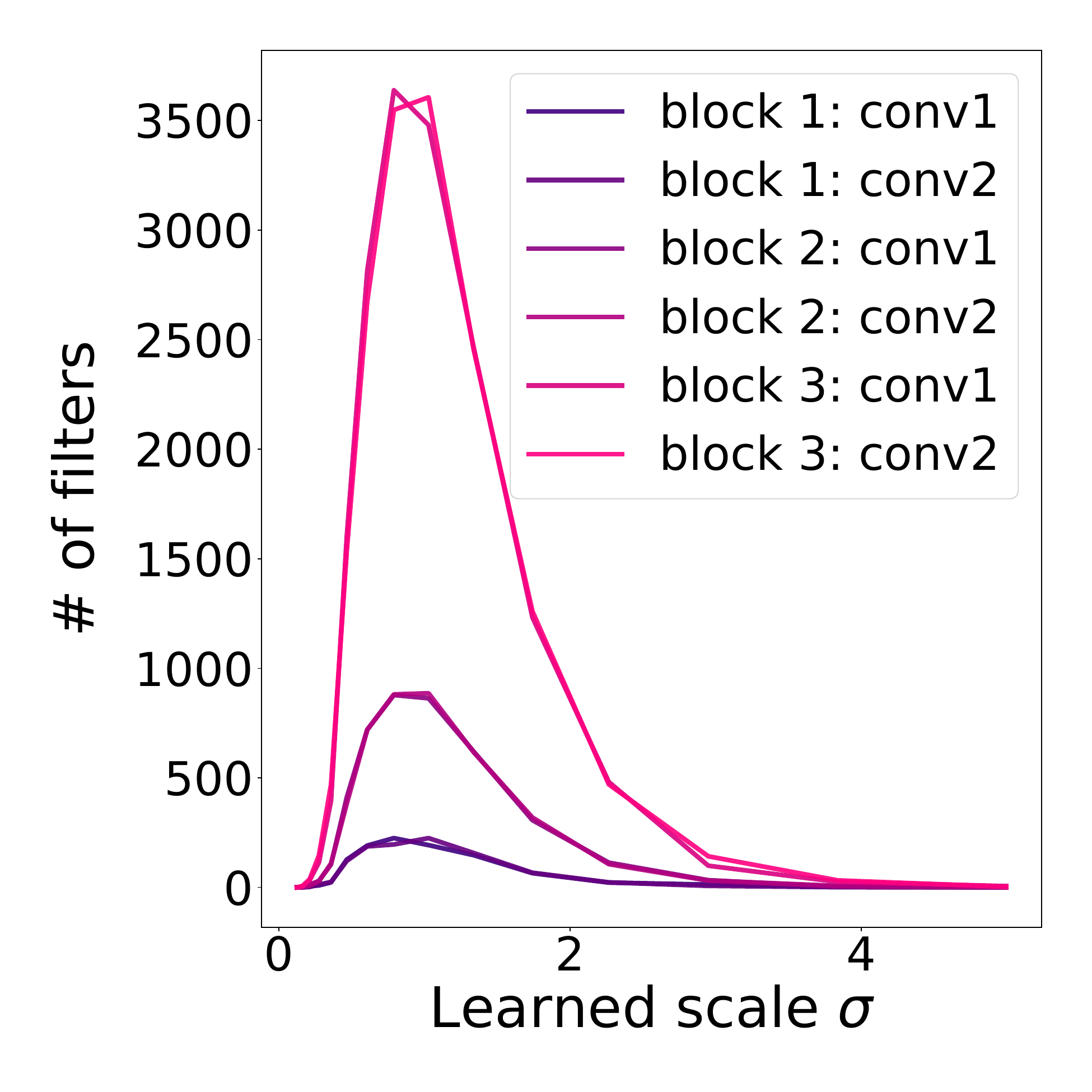}
        \end{tabular}
        &
        \hspace{0.7cm}
        \begin{tabular}{c}
            \includegraphics[trim=120 20 50 120, width=0.226\linewidth]{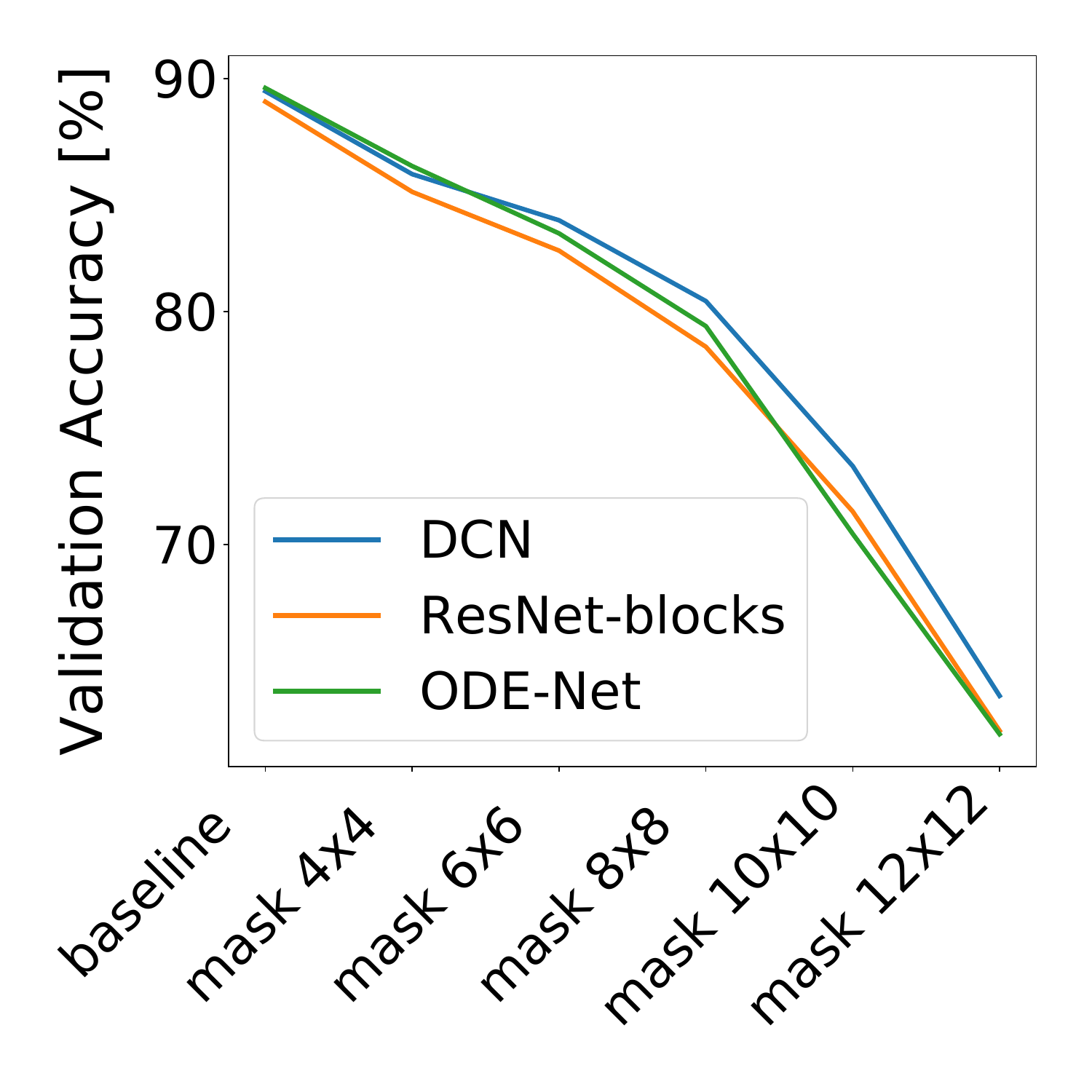}
        \end{tabular}    \\
        (a) & (b) & (c) \vspace{-0.5cm}
    \end{tabular}
    \caption{
        (a) Learned $\sigma$ values increase with depth within the network.
        (b) $\sigma^{ji}$ distributions within the ODE blocks display a positive skew in line with biological observations. (c) CIFAR-10 validation accuracies on the pattern completion task with increasing mask size.}
    \label{fig:sigmas}
\end{figure*}

\textbf{Scale fitting.} As an advantage over conventional CNNs, it is possible to directly investigate the optimal receptive field (RF) size in each \model~block after training, since \model s fit the kernel scale $\sigma$ explicitly.
We observe an upward trend in the SRF scale $\sigma$ with the depth of the convolutional layer within the network (Fig.~\ref{fig:sigmas}a). While the RF size grows with depth also in conventional CNNs, it typically grows in a predetermined manner: for a cascade of convolutional layers the RF size is a linear function of depth given constant kernel size. Thus, the receptive field size at every CNN layer is fixed depending on the architecture and hyperparameters. This is a limitation of CNNs which the visual system does not necessarily have. \model s, on the other hand, can \textit{learn} RF sizes which grow non-linearly as a function of depth, which seems to be in line with the behaviour in downstream visual areas~\citep{Smith2001}.

In addition, we plot the distribution of learned $\sigma^{ji}$ in different ODE blocks of the model \model~$\sigma^{ji}$ (Fig.~\ref{fig:sigmas}b). Note that the scale parameter $\sigma$ controls the bandwidth of the SRF filters and is thus related to their spatial frequency response. We find that the $\sigma^{ji}$ distributions after training are approximately log-normal and display a positive skew, which is consistent with the scale and spatial frequency tuning distributions in the primate visual system~\citep{Yu2010}. We believe these results are promising for bridging the gap between deep learning and traditional models of biological systems.

\textbf{Pattern completion.} 
Established models from computational neuroscience, with continuous temporal dynamics and well-defined recurrent interaction structures, such as the Ermentrout-Cowan model~\citep{Bressloff2001}, or neural field models~\citep{Amari1977}, display interesting high-level phenomena such as spontaneous pattern formation and travelling waves~\citep{Coombes2005}.
Such models employ local, distance-dependent interactions, similar to the SRF-based ODE blocks in the \model~formulation.
Based on this resemblance, we explore whether \model s may display similar \emph{emergent} properties. Specifically, we hypothesize that \model s can perform well in the case of locally missing information in images, through pattern completion at the feature map level.

\begin{table}
    \centering
    \begin{tabular}{lc}
        \toprule
        {Model} & {Parametrization} \\
        \midrule
        \model~$\sigma(t)$     & $\sigma=2^{at+b}$  \\
        \model~$\sigma(t^2)$         & $\sigma=2^{at^2+bt+c}$ \\
        \model~$\sigma(t)$, $\alpha(t)$ & $\sigma=2^{a_s t+b_s}$, $\alpha=a_{\alpha} t+b_{\alpha}$ \\
        \bottomrule
    \end{tabular}
    \caption{Meta-parametrization of filter parameters $\sigma$ and $\alpha$ as a function of depth $t$ in different \model~variants.}
    \label{tb:meta-def}
\end{table}
\begin{table}
    \begin{tabular}{lc}
        \toprule
        Model & Accuracy (\%) \\
        \midrule
        \model~$\sigma(t)$     & 89.97 $\pm$ 0.30 \\
        \model~$\sigma(t^2)$         & 89.93 $\pm$ 0.28 \\
        \model~$\sigma(t)$ and $\alpha(t)$ & 89.88 $\pm$ 0.25 \\
        \bottomrule
    \end{tabular}
    \caption{CIFAR-10 validation accuracies averaged over 3 runs for \model~models with meta-parametrization.}
    \label{tb:meta-accuracy}
\end{table}

\begin{figure}[t]
    \includegraphics[width=0.99\textwidth]{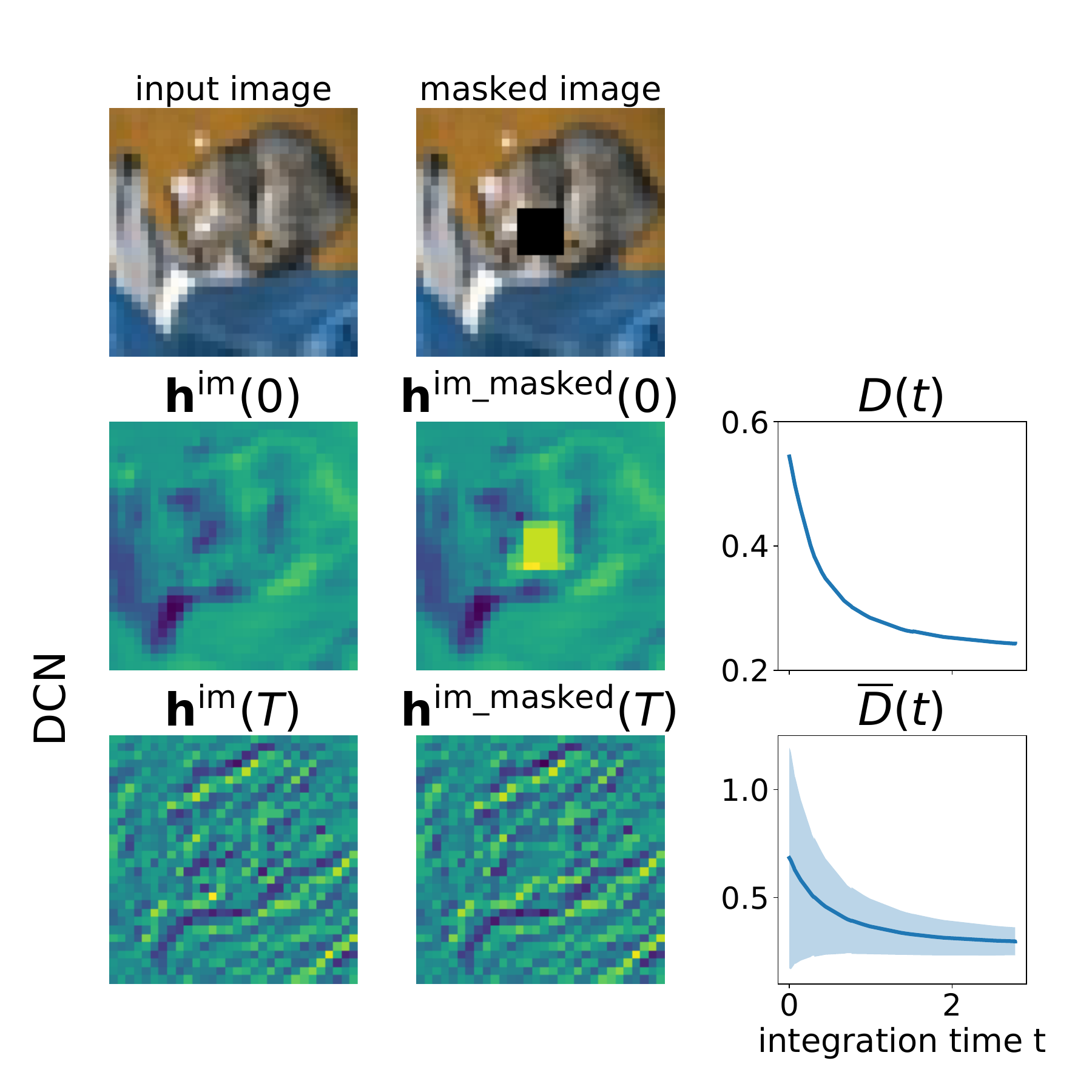}
    \vspace{-0.4cm}
    \caption{Pattern completion in the DCN feature maps during classification of masked images. 
    Feature maps in a single channel of ODE block 1 are shown for an example image. We find that the difference $D(t)$ between the feature maps $\textbf{h}^\textrm{im}(t)$ of an intact image and $\textbf{h}^\textrm{im\_masked}(t)$ of a masked image is reduced as $t \rightarrow T$. We also show the mean $\overline{D}(t)$ for 1000 validation images (bottom right), where the shaded area is the standard deviation over different images. Example feature maps from baseline models are provided in Appendix~A.5.\vspace{-0.3cm}}
    \label{fig:fm_evolution}
\end{figure}

We test this hypothesis on the \model~models trained on CIFAR-10 classification by masking $n \times n$ pixels of the validation images at test time. The masks have zero pixel values, and are placed at the center of the image. We find that when confronted with a small patch of missing information at test time, \model s can generate feature maps similar to those obtained from intact images.
Specifically, we observe that the total difference:
\begin{equation}
{D(t)=\frac{1}{A}\sum |\textbf{h}^\textrm{im}(t)-\textbf{h}^\textrm{im\_masked}(t)|}
\end{equation}
between the feature maps generated by an intact image $\textbf{h}^\textrm{im}(t)$ and a masked image $\textbf{h}^\textrm{im\_masked}(t)$, normalized by the amplitude of the intact image $A$, is reduced within an ODE block (Fig. ~\ref{fig:fm_evolution}).
In terms of the overall classification performance with images masked at test time, we find that \model s are marginally more robust against zero-masking than baselines (Fig.~\ref{fig:sigmas}c).

\subsection{Contrast robustness and computational efficiency}
\begin{figure}
    \includegraphics[width=0.9\textwidth]{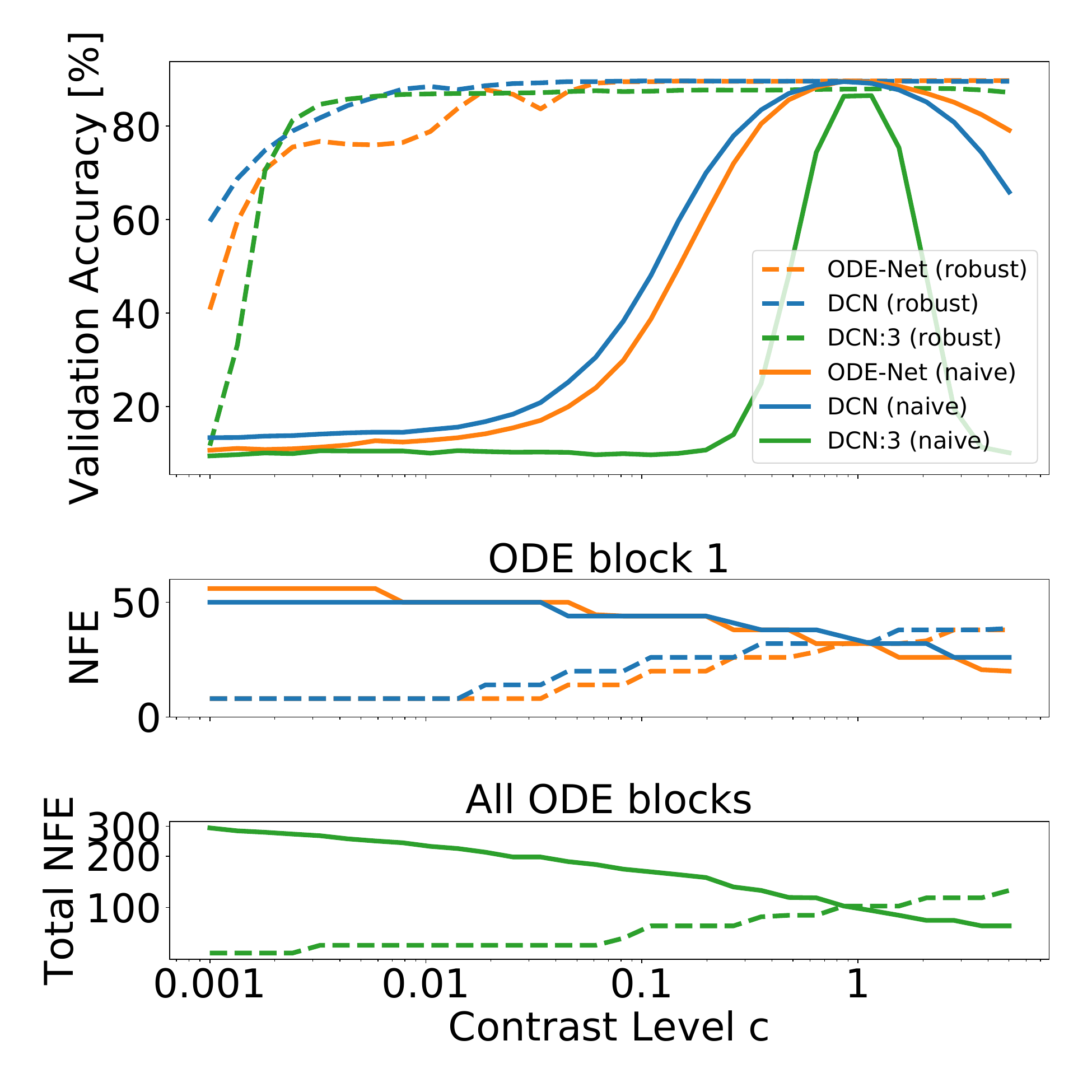}
    \vspace{-0.4cm}
    \caption{On the CIFAR-10 validation set, \model s are more robust than baseline ODE-Nets to changes in input contrast $c$ at test time (top). Interestingly, the number of function evaluations (NFEs) in the first ODE block (middle) or the whole \model~network (bottom) can be reduced considerably by modulating $c$.\vspace{-0.4cm}}
    \label{fig:contrast_rob}
\end{figure}

The selectivity of neuronal responses is invariant to contrast in mammalian vision~\citep{Sclar1982,Skottun1987}. However, we observe that \model~and ODE-Net models are sensitive to changes in input contrast. This is not unexpected since ODE blocks compute the solution to the initial value problem posed by the equations of motion and the input $\mathbf{h}(0)$. To quantify this sensitivity we vary the contrast $c$ of the input images at test time, where for each image $H$ in the CIFAR-10 validation set we define the network input as $\widehat{H}=cH$.
When naively changing the input contrast $c$ this way, we find that the validation accuracy decays rapidly for both models (solid lines in Fig.~\ref{fig:contrast_rob}, top).

Empirically, we notice that, with the appropriate choice of normalization functions, the input contrast $c$ has a direct effect on the time scales of the solution $\mathbf{h}(t)$. This means that under different contrast values $c$, the feature map trajectories within an ODE block may converge faster, and a more efficient \model~implementation might be possible.
Based on this observation, we heuristically test whether scaling the integration time interval $T$ (used during training) of ODE block 1 by the input contrast at test time, as $\widehat{T} = cT$, can improve contrast robustness at test time. We find that with the scaled integration interval, \model~validation accuracy is relatively robust against changes in contrast $c$, compared to naive baselines and ODE-Net, until $c<<1$ when time scales become too fast and the ODE solver becomes unstable for all models (dashed lines in Fig.~\ref{fig:contrast_rob}, top).

Interestingly, we observe a reduction in the number of function evaluations (NFEs) in ODE block~1 for $c<1$ (Fig.~\ref{fig:contrast_rob}, middle). Furthermore, we show that as long as the error tolerance of the ODE blocks are not decreased, this effect can be exploited by scaling the input feature maps of all ODE blocks by $c$ for significant computational savings. We find that decreasing $c$ leads to considerable efficiency improvements, where total NFEs can be reduced from 102 to 60, (for $c=1$ and 0.06), with less than 0.5\% loss in accuracy (Fig.~\ref{fig:contrast_rob}, bottom).

%% file: related.tex
\section{Related Work} \label{sec:related}
Our proposed \model~networks extend prior work on continuous filters and continuous depth neural ODEs.

\textbf{Spatially continuous filter representations.}
Structured filters have been traditionally used in computer vision for extracting image structure at multiple scales.
N-jet filter basis is first introduced by~\citet{florack1996gaussianLocalJet} based on previous work on Gaussian scale-spaces~\citep{florackIVC92scaleAndDiffStruct,lindebergBook13scaleSpaceInCV}.
We use the N-jet basis, which enables a spatially continuous representation, with a learnable scale parameter $\sigma$, to approximate convolutional filters.

Similar to the N-jet basis, a set of oriented multi-scale wavelets, called a steerable pyramid, is proposed by \citet{simoncelli1992shiftable} and complex wavelets have been used by~\citet{mallat2012groupInvScat} and~\citet{brunaPAMI13scatter} as part of scattering transforms.
CNN filters based on linear combinations of Gabor wavelets are adopted by \citet{luan2017gaborCNNs}, 
while \citet{WorrallCVPR17harmonicNets} propose circular harmonics, as spatially continuous filter representations.

Similar to our approach, \citet{shelhamer2019blurring} combine free-form filters with Gaussian kernels, thus learning the filter resolution.
Likewise, \citet{xiong2020variational} learn filter sizes using Gaussian kernels optimized using variational inference.
Finally, \citet{Loog2017} integrate continuous scale-selectivity through a regularization hyper-parameter.
Here, we use the N-jet framework based on Gaussian derivatives as in \citet{jacobsen2016structured} and \citet{Pintea2021}, however our main motivation is retaining compatibility with biological models. Also, unlike \citet{jacobsen2016structured} we learn the scale parameter $\sigma$ during training.

\textbf{Continuous depth representations in deep networks.}
Along with work by~\citet{Lu2018} and~\citet{Ruthotto2019}, networks continuous in the depth (or time) dimension have been proposed by~\citet{Chen2018} under the name neural ordinary differential equations (ODEs).
They propose ODE-Nets based on the ResNet formulation~\citep{He2015} for classification tasks, which we use as a baseline.
In this work we focus mainly on image classification, however, there is extensive ongoing work on generative models and normalizing flows using the neural ODE continuous depth interpretation~\citep{salman2018deep,grathwohl2019ffjord}. We note that \model s can be readily incorporated into continuous flow models, as well as other spatio-temporally continuous CNN interpretations based on partial differential equations~\citep{Ruthotto2019}.

Even though the adjoint method described in~\citet{Chen2018} offers considerable computational savings, especially in terms of memory, recent work has improved upon it both in terms of stability, computational efficiency and performance~\citep{dupont2019augmented,Finlay2020,Zhuang2020ODE}. Likewise, the contrast robust formulation of \model s, as well as the synergy between the  $\mathcal{O}(1)$ memory complexity of the adjoint method and spatially separable SRF filters (the implementations of which may otherwise inflate the memory cost) provide potential computational benefits over conventional CNNs where the number of function evaluations is fixed.

Other studies have suggested that, similar to our \model~variants where the filter definitions are independent of depth, neural ODEs based on ResNet architectures with weight sharing can be interpreted as recurrent neural networks~\citep{Kim2016,rousseau2019residual}, which bridges the gap between deep learning, dynamical systems and the primate visual cortex~\citep{Liao2016,Massaroli2020}.
Similar to these works, we illustrate the parallels between neural ODEs and the dynamical systems approach of the computational models of biological circuits. As a novel contribution, we extend neural ODEs to \model s, where not only the depth of the network is continuous but also the shape and spatial resolution of the filters are end-to-end trainable.

\textbf{CNNs and RNNs as models of biological networks.}
There is extensive prior work on CNNs and recurrent neural networks (RNNs) for modeling biological computation. The visual cortex is highly recurrent~\citep{Dayan2001,Liao2016} which is thought to be responsible for complex neuronal dynamics~\citep{BenYishai1995,Angelucci2006}.
Accordingly, computational models with lateral connections~\citep{Sompolinsky1988, Ernst2001} and more recently RNNs~\citep{Laje2013,Mante2013,Mastrogiuseppe2018} have been extensively used as models of biological neural computation. For example the first-order reduced and controlled error (FORCE) algorithm, have been used to reproduce the dynamics of different biological circuits~\citep{Sussillo2009,Laje2013,Carnevale2015,Rajan2016,Enel2016}. Similarly, optimization via gradient-based algorithms such as the Hessian-free method (HF) or stochastic gradient-descent (SGD) have been adopted to replicate experimental observations~\citep{Mante2013,Barak2013,Song2016}. It has also been suggested to use spiking recurrent networks~\citep{Kim2018,Kim2019} and incorporate synaptic dynamics~\citep{Ba2016,Miconi2018} for improved physiological realism.

Bringing together the power of CNNs and neuroscience, recurrent convolutional networks (RCNNs) have been proposed~\citep{Liang2015,Spoerer2017,Hu2018}, which can emulate biological lateral connectivity structures and extra-classical receptive field effects. Similar to our work where depthwise-continuity mimics recurrent networks, it has been shown that adding recurrent layers to convolutional deep networks can facilitate pattern completion in a manner consistent with psychophysical and electrophysiological experiments~\cite{Tang2018}. Furthermore, our \model~models have the potential to compress the depth of the network, by replacing multiple sequential layers with meta-parametrized ODE blocks, which are analogous to recurrent networks with continuously evolving filter parameters. In a similar line of work, it has previously been shown that shallow networks with recurrently connected layers can achieve high object recognition performance while retaining brain-like representations, and specifically reproducing the population dynamics in area IT of the visual system much more closely than feed-forward deep CNNs~\cite{Kar2019,Kubilius2019}.

In contrast to standard RNNs, our model is based on the ResNet inspired model of neural ODEs, and in its current form (Eq.~\ref{eq:EoM}), does not accept time-variant input. In that sense, the spatio-temporal dynamics of \model s refer to the dynamics of the feature maps, or neuronal responses, and not the input. Nevertheless, this gives \model s the ability to model time-varying responses, even to static input images.

In addition, \model s with weight sharing can be thought of as recurrent networks~\citep{rousseau2019residual} and can be easily modified to process time-variant input (such as videos). However, in this paper we consider \model~models as an extension of conventional feed-forward CNNs, with extended temporal dynamics and continuous spatial representations, which are applicable to feed-forward models of the visual system similar to works by~\citet{BrainScore,Lindsey2019,Ecker2019,Zhuang2020}.

%% file: discussion.tex
\section{Discussion}

We introduce \model s, CNN models which learn spatio-temporally continuous representations, consistent with biological models.
We show that \model s can match baseline performance in an image classification task and outperform baselines in the small-data regime and in a reconstruction task, while using a smaller number of parameters. Similarly, we propose different methods of meta-parametrization of the convolutional filter as a function of depth, which may not only be applicable to network compression, but also for modelling the temporal profiles of biological responses.
As a further link with biological models, we have demonstrated that the learned filter scale distributions in \model s are compatible with experimental observations. This makes the \model~models viable for future neuroscientific investigations regarding the emergence of RF sizes. In addition, we have presented the capability of \model s to reduce errors in feature maps caused by masking. Finally, we have empirically shown an interesting interplay between the input contrast to ODE blocks and the time scales of the solutions, which can be capitalized on for computational savings.

However, one of the biggest limitations of \model~models is that they may become unstable during training. Combining neural ODEs with scale fitting may lead to exploding filter sizes at large learning rates. Especially for meta-parametrization, it would be advisable to clip the integration time and filter parameters within a reasonable range.

Nevertheless, we believe there are exciting future research opportunities involving \model s.
Neural ODE formulations provide an interesting opportunity for establishing a theoretical understanding of deep networks based on dynamical systems. The interplay of input contrast and integration time is one such observation which requires further investigation. Similarly, our choice of filters based on well-behaved Gaussian derivatives allow for further analytical studies, unlike conventional CNNs.

Similarly, \model s offer interesting possibilities for biological modelling. The inbuilt smooth evolution of filters in \model s can be used, for example, to incorporate response dynamics such as synaptic depression or short-term potentiation~\cite{Ba2016,Miconi2018}. Likewise, the equations of motion can be modified to reflect axonal delays or generate oscillations.
Taken together, we believe by offering a link between dynamical systems, biological models and CNNs, \model s display an interesting potential to bring together ideas from both fields.

\section*{Acknowledgements}
\small Authors thank the reviewers for insightful comments, and Prof.~Dr.~Marco~Loog for fruitful discussions. This publication is part of the project ``Pixel-free deep learning" (TOP grant with project number 612.001.805) which is financed by the Dutch Research Council (NWO).

%% file: appendix.tex
\appendix
\section{Appendix}

\subsection{Gaussian multiscale local N-jet} \label{ap:njet}
Based on the Schwartz theory of smooth test functions, the Gaussian scale-space paradigm states that the derivatives $L_{i_1 \ldots i_n}(\vec{x};\sigma)$ of a function $L_0(\vec{x})$ with respect to the spatial variables $x_{i}$, with $i=1\ldots d$, and at scale $\sigma$ is given by the convolution
\begin{equation} \label{eq:ap_srf}
    L_{i_1 \ldots i_n}(\vec{x};\sigma) = L_0 \ast \partial_{i_1 \ldots i_n} G(\vec{x};\sigma)
\end{equation}
where $\partial_{i_1 \ldots i_n}$ is the $n$-th order partial derivative and $G(\vec{x};\sigma)$ is the normalized, isotropic Gaussian kernel with standard deviation $\sigma_i=\sigma$ and mean ${\mu_{i}=0}$~\citep{florack1996gaussianLocalJet}. Note that $L_0(\vec{x})$ does not need to be a smooth function, and therefore the Gaussian scale-space paradigm can be applied to obtain local image derivatives in different scales, where $\vec{x}$ denote the spatial coordinates and $\sigma$ can be interpreted as the coordinate in the scale dimension.

We build upon this definition of local image derivatives to build our structured receptive fields (SRFs), similar to~\citet{jacobsen2016structured}. The main idea we leverage is that using a Taylor approximation, one can decompose an input image into a superimposition of its local derivatives. This decomposition can then be performed by local convolution kernels in CNNs, where the relative weight of each derivative order can be learned during training. In order to show this, we observe that the $N$-th order Taylor expansion of an image $L: \mathbb{R}^2 \rightarrow \mathbb{R}$ around a point $(a,b)$ is given by
\begin{align}\label{eq:ap_taylor}
L(x,y) = \sum^N_{l=0} \sum^{N-l}_{k=0} \frac{(x - a)^l(y - b)^k}{l! k!} \frac{\partial^{l+k}}{\partial x^l \partial y^k} L(a,b)
\end{align}
where the partial derivatives of $L$ with respect to $x$ and $y$ can be interpreted as $L_{1_1 \ldots 1_n}(x,y;\sigma_0)$ and $L_{2_1 \ldots 2_n}(x,y;\sigma_0)$ from equation~\ref{eq:ap_srf} at some original scale $\sigma_0$. This means that, via the linearity of convolution, and under the assumption that $L(x,y;\sigma)$ does not diverge, it is equivalent to either use the $N$-th order derivatives of the input image, or use the $N$-th order derivatives of the Gaussian function $G(\vec{x};\sigma)$, to perform the image decomposition at scale $\sigma$. The local N-jets can thus be seen to be parametrized by the coefficients in the expansion given in equation~\ref{eq:ap_taylor}. By definition, we consider the Taylor expansion coefficients to be covariant derivatives of the image $L(x,y;\sigma)$, which are independent of the local coordinate system, to reach the filter approximations given in equation~1 in the paper, where the coefficients take the form $\alpha_{l,k}$.

In short, the multiscale local N-jet provides a natural decomposition of an image in a local neighborhood in the spatial and scale dimensions. Under convolution with $G(\vec{x};\sigma)$, this decomposition provides a framework for defining convolutional filters in a CNN using $N$-th order Taylor polynomials.
The SRF filters we use are based on this N-jet definition and allow us to learn the scale, spatial frequency and orientation of the filters during training, which are fundamental properties of biological receptive fields~\citep{Jones1987,Lindeberg1993}.

In addition, the Gaussian scale-space formulation of SRFs~\citep{jacobsen2016structured} lead to theoretically interesting properties which strengthen the motivation for our choice of filters based on the Gaussian N-jet. For example, the semi-group property of Gaussian scale-spaces indicates for a Gaussian derivative kernel $G^{l,k}(x,y;t)$ parametrized by the variance $t=\sigma^2$
\begin{equation}
    G^{l,k}(x,y;t+t')=G^{l,k}(x,y;t) \ast G(x,y;t')
\end{equation}
where the superscripts $(l,k)$ denote the partial derivatives with respect to $x$ and $y$. This means that a translation in scale dimension can be simply achieved through convolution with a (0-th order) Gaussian kernel $G$. For \model s, this means that we have a solid understanding of the scale of the feature maps (in the sense of the Gaussian scale-space) at every layer $m$ in the network, as we know the value of $\sigma_m$ after training. In the absence of SRFs with an explicit scale parameter, this information is lost.

Similarly, SRF filters based on Gaussian derivatives are steerable by the coefficients $a_{l,k}$. For example, a second order filter with orientation $\theta$ can be described as a sum of basis functions
\begin{align}
    G^{2,0}_{\theta} & =a_{2,0}G^{2,0}+a_{1,1}G^{1,1}+a_{0,2}G^{0,2} \\ \nonumber
    &= \cos^2(\theta)G^{2,0}-2\cos(\theta)\sin(\theta)G^{1,1}+\sin^2(\theta)G^{0,2}.
\end{align}

Finally, the set of Gaussian derivative basis functions $G^{l,k}(x,y;\sigma)$ are spatially separable
\begin{equation}
    G^{l,k}(x,y;\sigma)=G^{l}(x;\sigma) G^{k}(y;\sigma)
\end{equation}
which is useful for computational efficiency in numerical applications.

\subsection{Training procedure} \label{ap:training}
The basic architecture of all our models is given in figure~2 of the paper. Unless otherwise stated, in all models we use group normalization~\citep{Wu2018} with 32 groups in every layer as the normalization function. As the nonlinear activation, we use continuously differentiable exponential linear units~\citep{Barron2017}, or CELU for \model~models. Based on the original ODE-Net implementation~\citep{Chen2018}, for {ODE-Net} models and ResNet-block models, we use rectified linear units (ReLU).
Likewise, when defining the integration time interval $T$, we stick to the original implementation, with $T=1$ for ODE-Net models, whereas we use $T=2$ for \model~models. Otherwise, all the hyperparameters are kept constant between models. For a brief overview of hyperparameter optimization, see appendix~\ref{ap:design}.

All models are trained for 100 epochs on the standard CIFAR-10 training set~\citep{cifar10} using cross-entropy loss. As data augmentation, we use random translations up to 4 pixels in each spatial dimension and random horizontal flips. Optimization is performed using SGD with a mini-batch size of 128, initial learning rate $10^{-1}$, momentum 0.9, and a learning rate decay by a factor of 0.1 at epochs 40 and 70.
For continuous time models based on neural ODEs, we use the adjoint method for backpropagating the losses and ODE solvers with error tolerance set to $10^{-3}$.

For convolutional layers with pixel-based $k \times k$ filters, the weights are initialized using the standard Kaiming uniform initialization~\citep{He2015init}. For layers using SRF filters, the initial $\alpha$ values are randomly sampled from a normal distribution with mean 0 and standard deviation 0.1, and initial $\sigma$ values are sampled from a normal distribution with mean 0 and standard deviation 2/3.

For the restricted CIFAR-10 experiments (small-data regime), we pick the total number of training images to be a multiple of our mini-batch size of 128, or otherwise have a minimal number of samples in the final batch (which is dropped in each epoch). For comparisons with the baseline results from~\citet{Arora2020}, we used exactly the same number of training images as in Table 2 of~\citet{Arora2020}. In order to confirm convergence, for the training set sizes $[520,1030,5120]$, we increased the number of training epochs by a factor of $[10,5,2]$ respectively.

For meta-parametrized models the initial values for the learnable parameters follow normal distributions $\mathcal{N}(\mu,\sigma_\mathcal{N})$: $a\sim\mathcal{N}(0,2/3)$, $b\sim\mathcal{N}(0,0.1)$, for the \model~$\sigma(t)$ model;
$a\sim\mathcal{N}(0,2/3)$, $b\sim\mathcal{N}(0,2/3)$, $c\sim\mathcal{N}(0,0.1)$ for the \model~$\sigma(t^2)$ model; and $a_s\sim\mathcal{N}(0,2/3)$, $b_s\sim\mathcal{N}(0,0.1)$, $a_{\alpha}\sim\mathcal{N}(0,0.1)$ and $b_{\alpha}\sim\mathcal{N}(0,0.05)$ for the \model~$\sigma(t)$, $\alpha(t)$ model.

For all our models using SRF filters, except for the \model~$\sigma^{ji}$ model, we use scale sharing within a convolutional layer such that $\sigma_m^{ji}=\sigma_m$ for all convolutional layers $m$, with input channel $i$ and output channel $j$. This makes the GPU implementation trivial, as all filters within a layer are sampled in the interval $[-2\sigma_m,2\sigma_m]$, and hence the convolutional kernel sizes within a layer are uniform. However, for the \model~$\sigma^{ji}$~model, a GPU implementation would be highly inefficient if we truncated the kernels at a fixed factor of $\sigma^{ji}$, independently for each input and output channel $i,j$. Therefore for the \model~$\sigma^{ji}$ model we fix the kernel size at $7 \times 7$, but we still learn the shape and the scale (bandwidth) of the filters during training.

We use the Dormand--Prince (DOPRI) method~\citep{DOPRI} as the numerical ODE solver. The DOPRI method is an explicit, adaptive solver of the Runge-Kutta family, which uses 6 function evaluations to compute fourth- and fifth-order accurate solutions to ODEs, along with an error estimate. The size of the adaptive steps taken by the solver can be regulated by specifying an error tolerance on this error estimate.

As is the case with most modern ODE solvers, the GPU implementation of the DOPRI solver that we use~\citep{Chen2018} considers the input arguments for the integration time interval $t \in [0,T]$ (or $t \in [0,2]$ in the case of all \model~models) as soft bounds. This means that the DOPRI algorithm may explore time points outside of this interval, based on its internal error estimation, and may then employ interpolation to return solutions within the specified bounds. In the meta-parametrized models, where for example $\sigma$ is an explicit function of $t$, this may lead to very large or very small kernel sizes, ordinarily unexpected within the integration time interval. In order to avoid numerical instability and memory issues in the meta-parametrized models, we scale down and clip the integration time $t$ when passing it into the parameter definitions as $\sigma(\tau t_\textrm{clip})$ and $\alpha(\tau t_\textrm{clip})$. We clip the $t$ values in the interval $[-0.5, 2.5]$ and use $\tau=0.5$.

\subsection{Hyperparameter tuning} \label{ap:design}
As mentioned in appendix~\ref{ap:training}, we share all the design choices and hyperparameters between all \model~and baseline ODE-Net models, except for the nonlinear activation function and integration time interval $T$.

\begin{table}[ht]
\captionsetup{width=0.5\linewidth}
\caption{CIFAR-10 validation accuracy (averaged over 3 runs) in the control experiments testing the effect of \model~model design choices on the baseline ODE-Net.}
\label{tb:ODE_hyperparameters}
    \begin{center}
        \begin{tabular}{lc}
            \toprule
            {Model} &  {Accuracy (\%)}\\
            \midrule
            \model-ODE & 89.46 $\pm$ 0.16 \\
            ODE-Net (baseline) & 89.60 $\pm$ 0.28 \\
            ODE-Net with $T=2$ & 89.50 $\pm$ 0.07 \\
            ODE-Net with CELU & 89.33 $\pm$ 0.16 \\
            ODE-Net with CELU and $T=2$ & 89.25 $\pm$ 0.30 \\
            \bottomrule
        \end{tabular}
    \end{center}
\end{table}

This difference in design choices arises since for the ODE-Net baseline we stay faithful to the original ODE-Net implementation, where ReLU is the nonlinear activation function and $T=1$, whereas for \model~models we use CELU as the activation function, due to its generalized compatibility with the adjoint method, and $T=2$.
In order to verify that our design choices do not provide an unfair advantage over the ODE-Net baseline, we run some control experiments, where we vary the activation function and $T$ in the ODE-Net baseline.

We find that the change of activation functions or integration interval $T$ do not provide a significant increase to the CIFAR-10 classification performance in the ODE-Net baseline (Table~\ref{tb:ODE_hyperparameters}).

\subsection{CIFAR-10 image reconstruction} \label{ap:reconstruction}
For the reconstruction task, we use an encoder with 3 \model-ODE blocks and a decoder with 3 \model-ODE blocks (as shown in figure 2 of the paper). For the baseline networks we replace the 3 \model-ODE blocks with ODE-Net or ResNet-blocks. The encoder is pre-trained on the CIFAR-10 classification task. We use the feature maps at the end of ODE Block 3 as the input to the decoder network. The decoder \model~network is made up of 2 ODE blocks separated by bilinear upscaling, $1 \times 1$ convolutions to reduce the number of channels, normalization, non-linear activation and a final $1\times 1$ convolution to generate the output in RGB space.

We implement reconstruction as a regression task and use the mean squred error (MSE) as the loss function. Otherwise, the training parameters are the same as the classification experiments: We use the SGD optimizer with a mini-batch size of 128, learning rate $10^{-1}$ and momentum 0.9, together with the adjoint method and an error tolerance of $10^{-3}$.

\begin{figure}
    \centering
    \includegraphics[trim=0 0 0 70, width=0.65\textwidth]{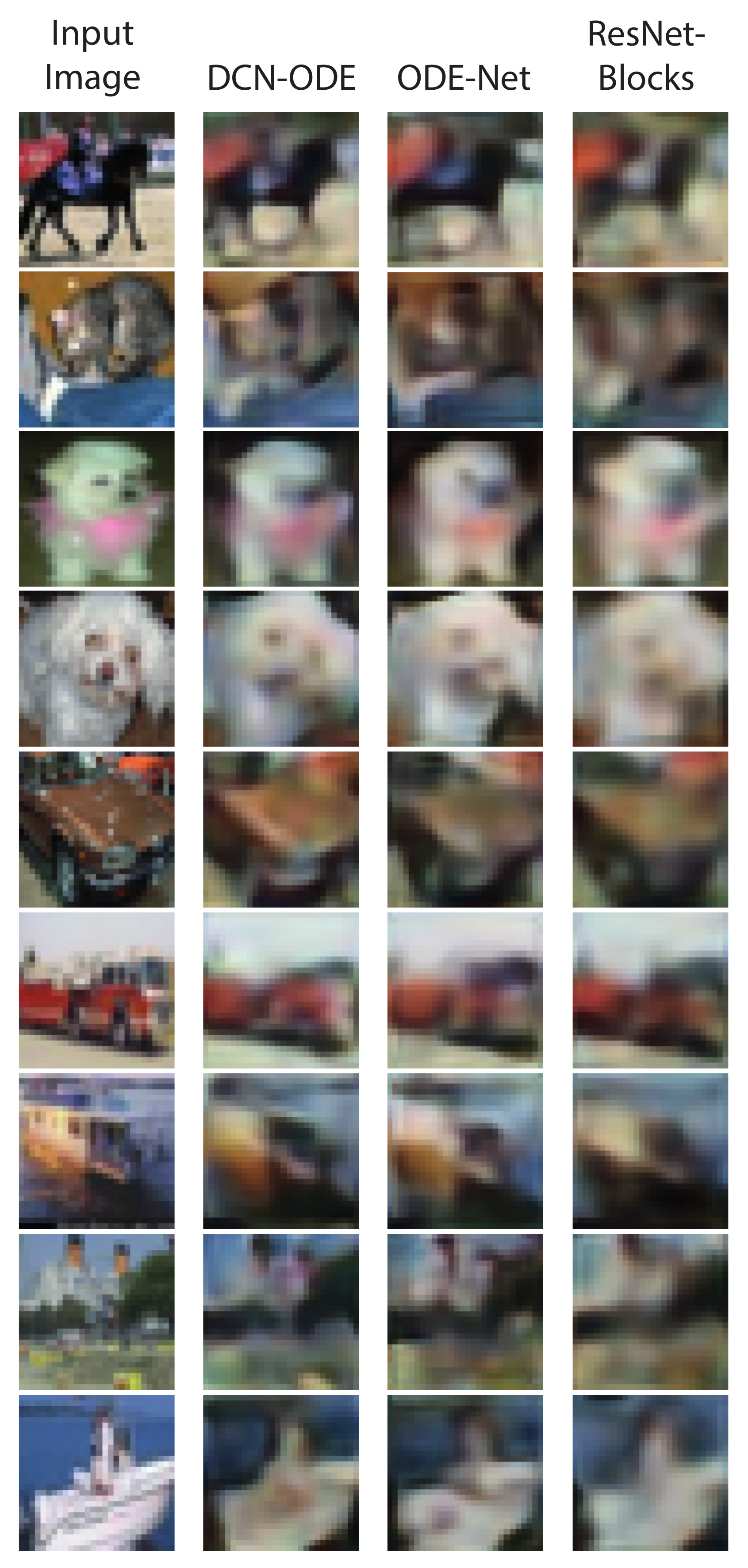}
    \caption{Example CIFAR-10 validation images and their reconstruction by the \model-ODE model as compared to baseline models.}
    \label{ap:recon_ims}
\end{figure}

Some example image reconstructions (randomly selected) from the CIFAR-10 validation set by the \model~and baseline networks are shown in figure~\ref{ap:recon_ims}.

\subsection{Pattern completion in feature maps} \label{ap:feature_maps}
\begin{figure}
    \centering
    \begin{tabular}{cc}
        \includegraphics[width=0.49\textwidth]{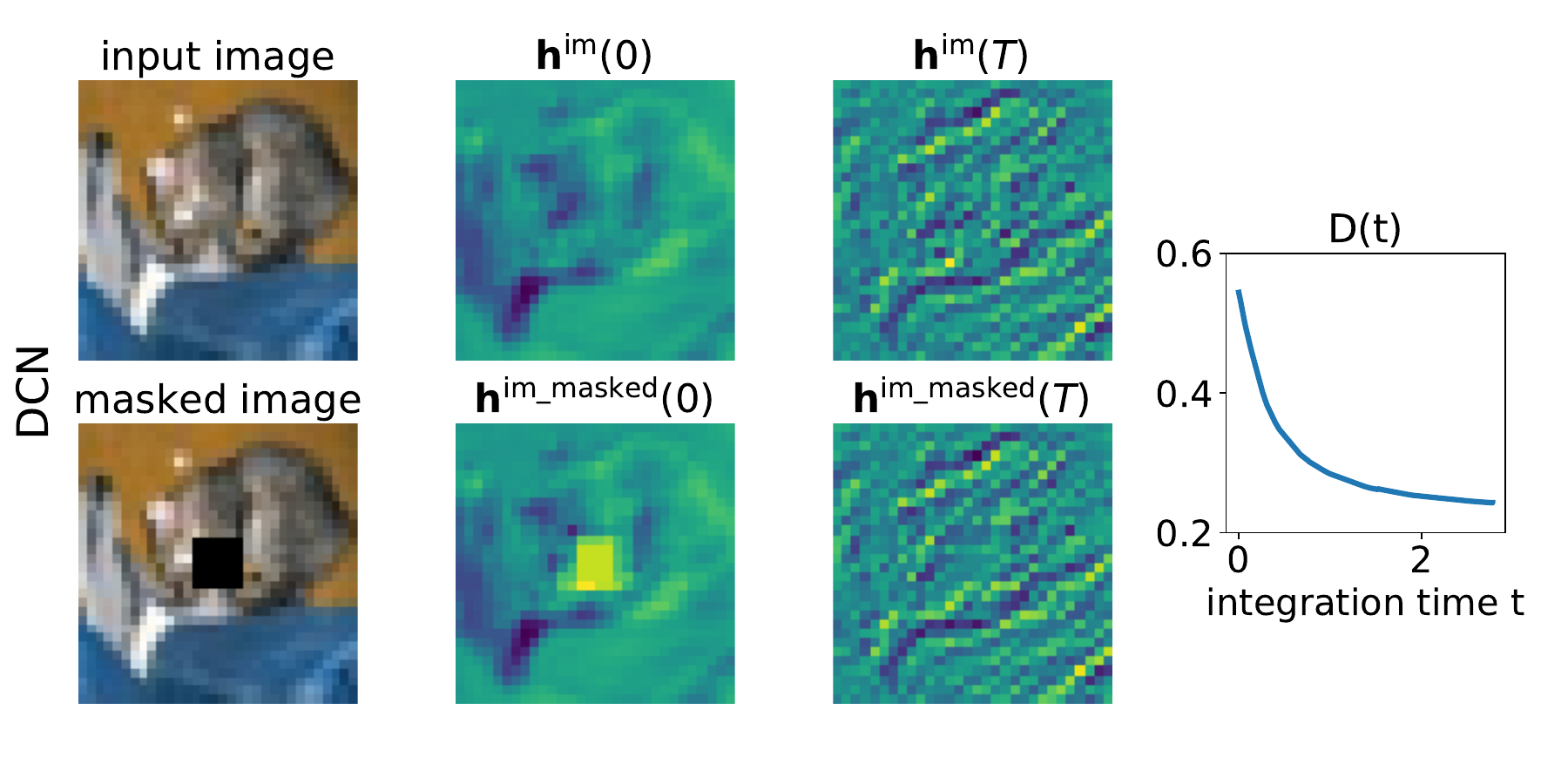}
         & 
         \includegraphics[width=0.49\textwidth]{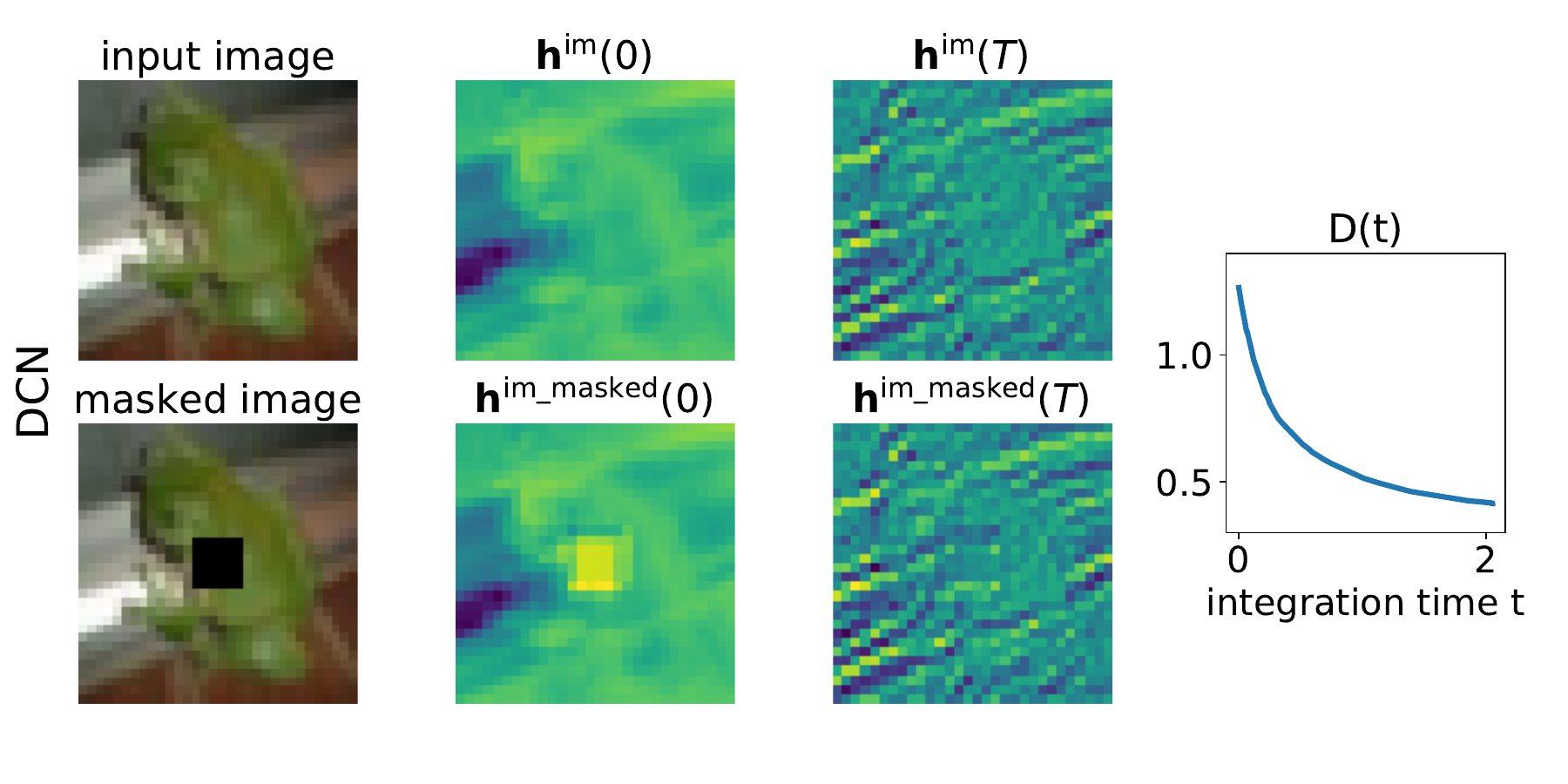}\\
         \includegraphics[width=0.49\textwidth]{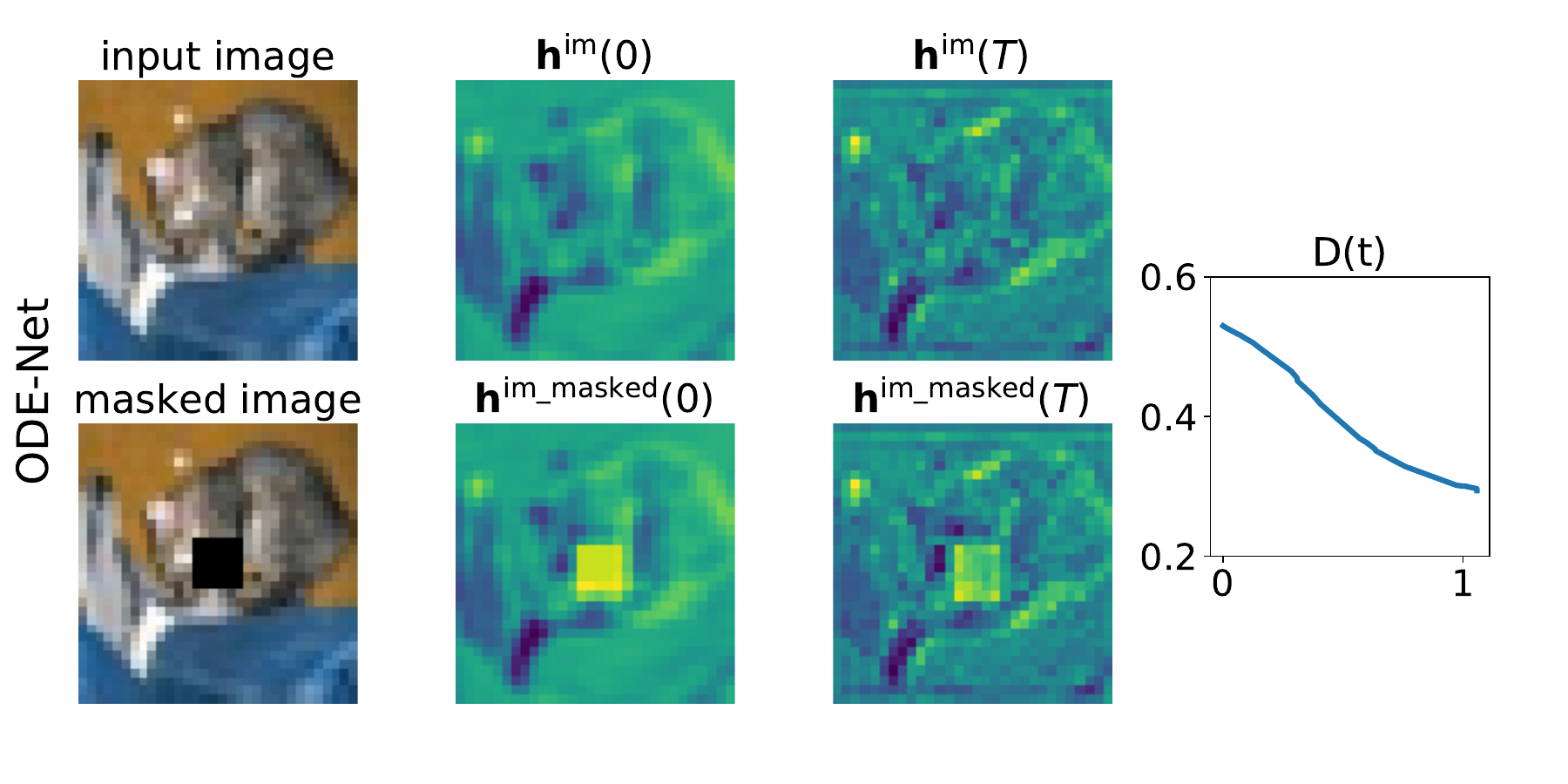}
         & 
         \includegraphics[width=0.49\textwidth]{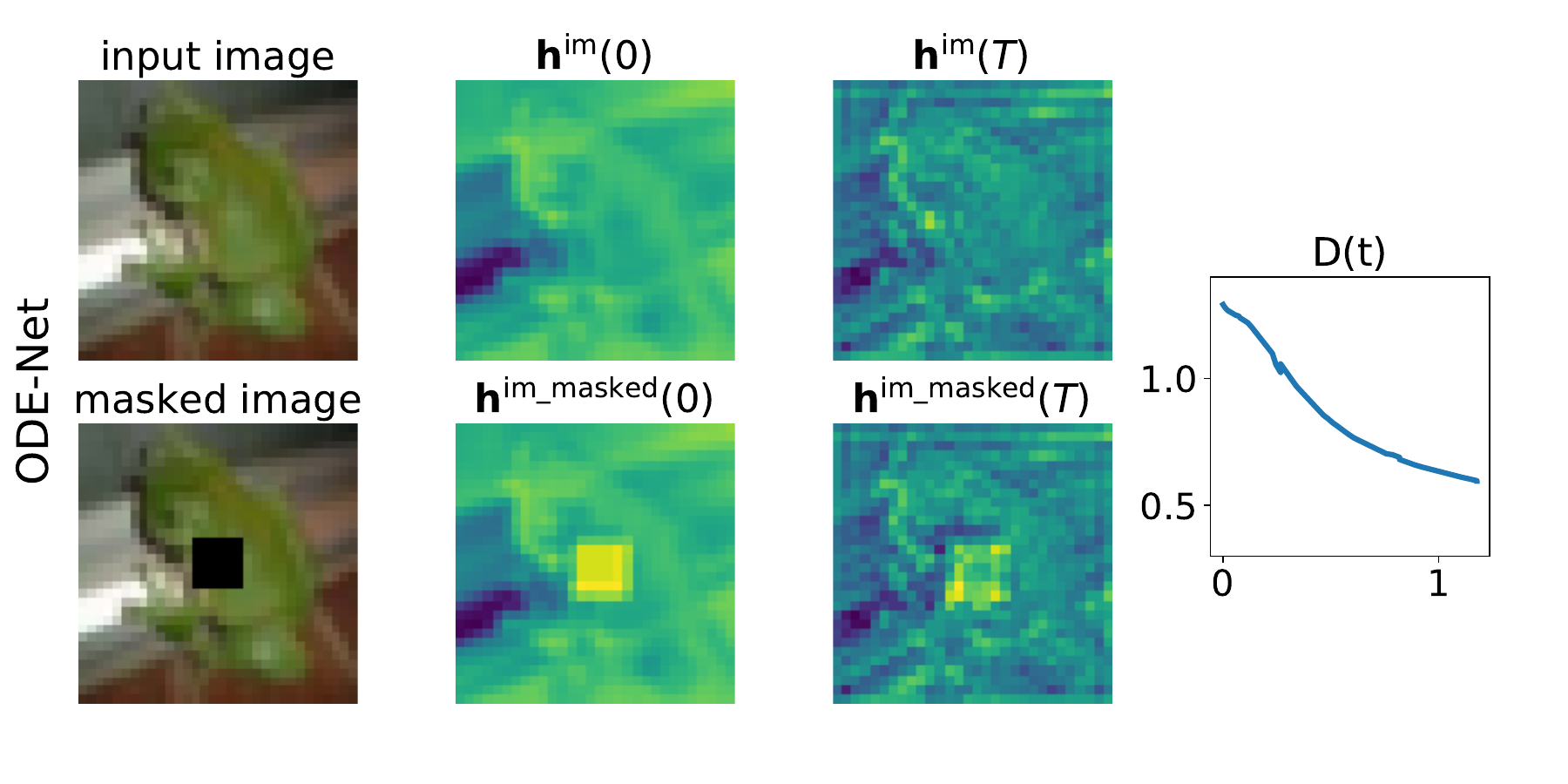}\\
         \includegraphics[width=0.49\textwidth]{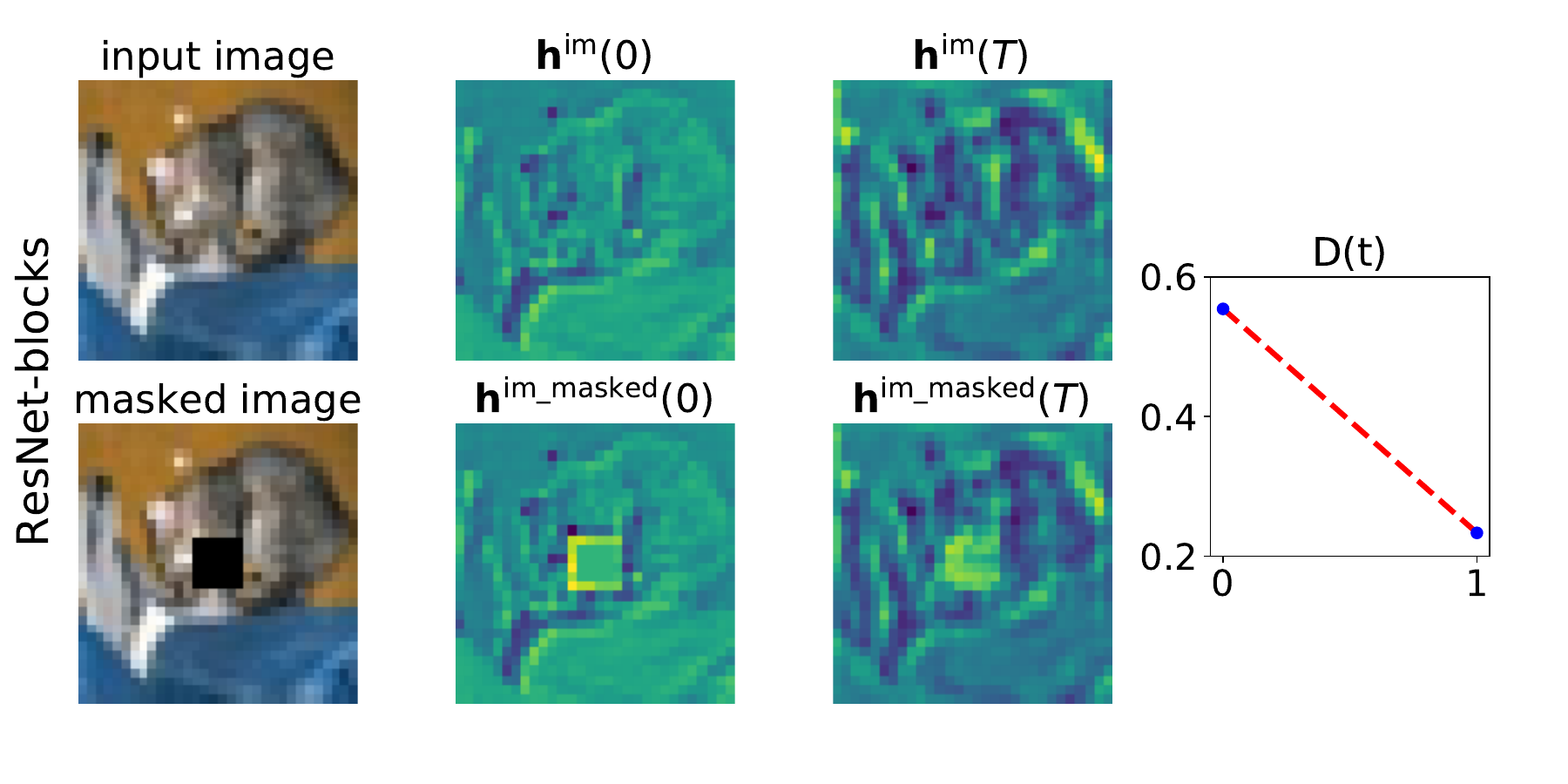}
         &
         \includegraphics[width=0.49\textwidth]{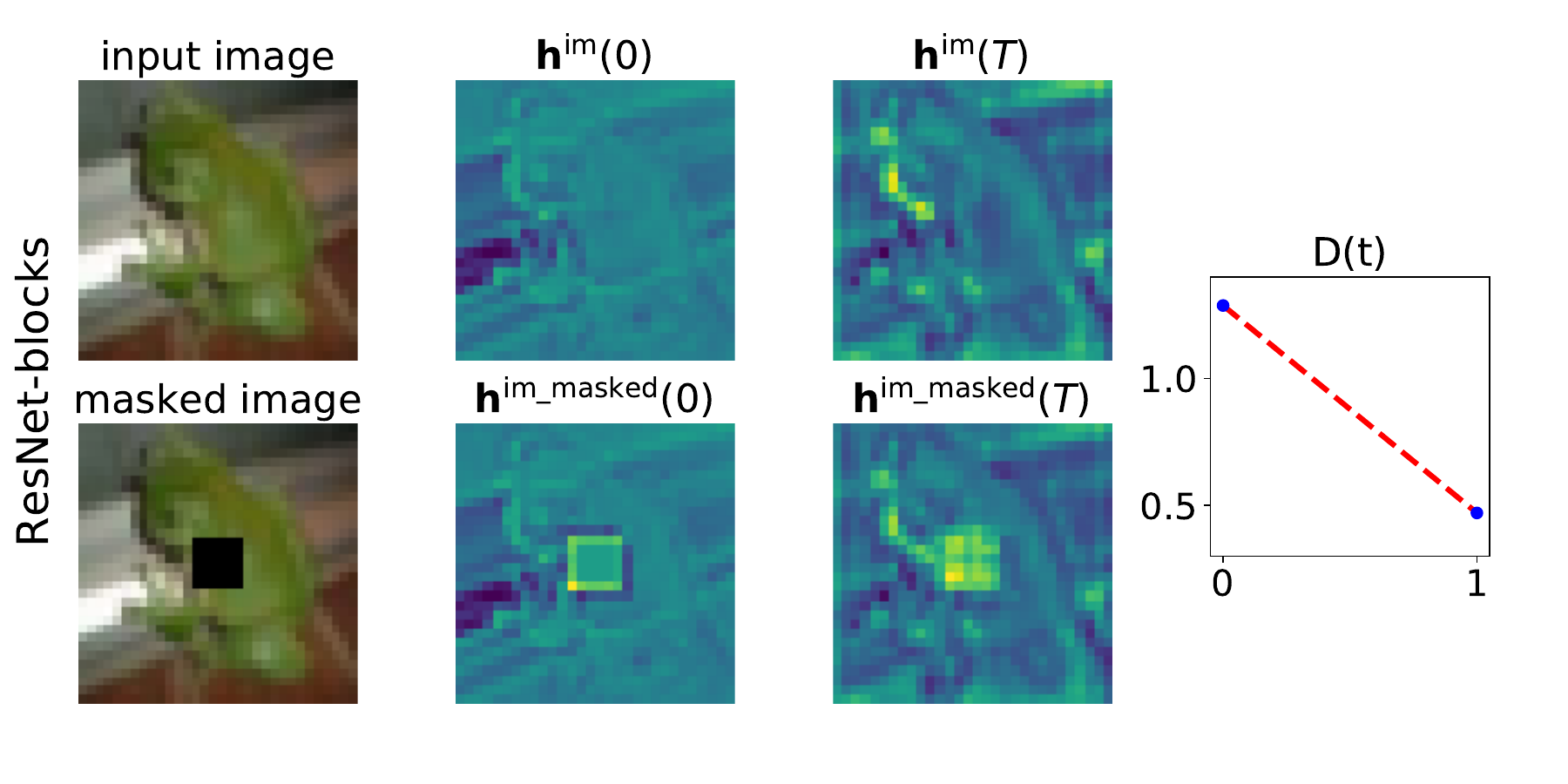}
    \end{tabular}
    \caption{Feature map evolution within the first ODE block (or ResNet block) of different models.}
    \label{fig:fm_evolution_ap}
\end{figure}

In figure~\ref{fig:fm_evolution_ap}, we show the feature map evolution within the first ODE block (or ResNet block) of different models with and without masking of some example input images. Size of the mask depicted here is $6 \times 6$ pixels and the example images were chosen so as to have the mask located close to the middle of the object. We picked some channels with visible mask-related artifacts in the input feature maps to the first ODE (ResNet) block. Since there is no feature map trajectory in the ResNet-blocks model, but only one input and one output feature map, the difference between the feature maps of the intact and masked image is given as a scatter plot of two data points connected by a red line.

\begin{figure}
    \centering
    \setlength\tabcolsep{3pt} 
    \begin{tabular}{ccc}
        \includegraphics[width=0.3\textwidth]{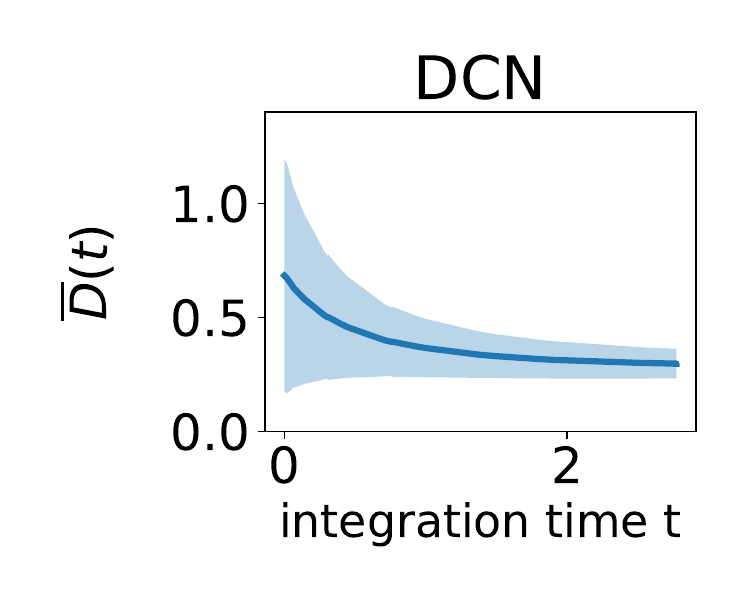}
         & \hspace{0.2cm}
         \includegraphics[width=0.3\textwidth]{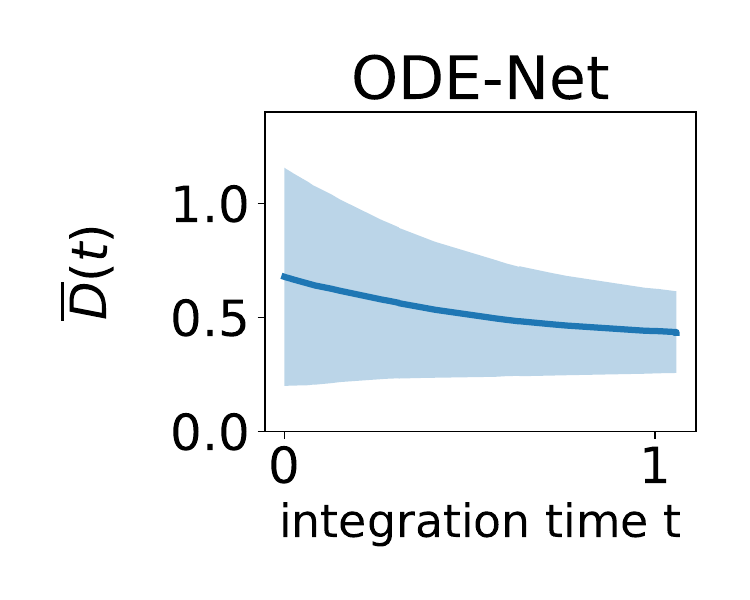}
         & \hspace{0.2cm}
         \includegraphics[width=0.3\textwidth]{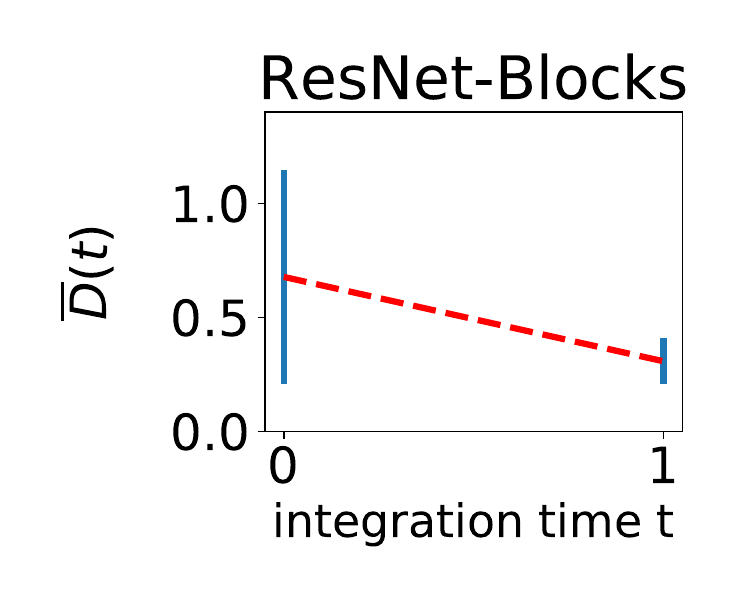}
    \end{tabular}
    \caption{Evolution of the mean difference $\overline{D}(t)$ between feature maps of an intact input image and a masked input image, averaged over 1000 images in the CIFAR-10 validation set. The shaded areas (or in the case of ResNet, the errorbars) show the standard deviation.}
    \label{fig:fm_evolution_mean}
\end{figure}

Figure~\ref{fig:fm_evolution_mean} depicts the average difference of intact and masked feature maps $\overline{D}(t)$ averaged over 1000 images and the standard deviation for the \model~and baseline networks.

